\documentclass[preprint,12pt,authoryear]{elsarticle}

\usepackage{amssymb}
\usepackage{amsmath}

\usepackage[utf8]{inputenc}
\usepackage[english]{babel}
\usepackage{graphicx}
\usepackage{booktabs}
\usepackage{multirow}
\usepackage{array}
\usepackage{xcolor}
\usepackage{hyperref}
\usepackage{cleveref}
\usepackage{listings}
\usepackage{algorithm}
\usepackage{algorithmicx}
\usepackage{caption}
\usepackage{subcaption}
\usepackage{enumitem}
\usepackage{longtable}
\usepackage[table]{xcolor}

\usepackage[margin=1in]{geometry}
\hypersetup{
    colorlinks=true,
    linkcolor=blue,
    citecolor=blue,
    urlcolor=blue
}

\journal{Computers \And Education}

\begin{document}

\begin{frontmatter}



\title{Automated grading of Linux/bash examinations using large language models: a four-level cognitive taxonomy approach} 


\author[inst1,inst2]{Manuel Alonso-Carracedo\corref{cor1}} 
\ead{manuel.alonso.carracedo@uvigo.gal} \cortext[cor1]{Corresponding author.}
\author[inst1,inst2]{Ruben Fernandez-Boullon} 
\ead{ruben.fernandez.boullon@uvigo.gal}
\author[inst1,inst2]{Pedro Celard} 
\ead{pedro.celard.perez@uvigo.gal}
\author[inst1,inst2]{Francisco J. Rodríguez-Martínez} 
\ead{ franjrm@uvigo.gal}
\author[inst1,inst2]{Lorena Otero-Cerdeira} 
\ead{locerdeira@uvigo.gal}

\affiliation[inst1]{organization={Universidade de Vigo, Department of Computer Science, ESEI-Higher School of Computer Engineering}, 
addressline={Edificio Politécnico, Campus Universitario As Lagoas s/n}, 
city={Ourense}, 
postcode={32004}, 
country={Spain}}

\affiliation[inst2]{organization={IFCAE-Institute for Research in Physics, Computing and Aerospace Science},
addressline={Universidade de Vigo}, 
city={Ourense}, 
postcode={32004}, 
country={Spain}}

\begin{abstract}

Scalable and reliable grading of command-line examinations remains a challenge in computing education, where rising enrolments make manual marking difficult and rule-based autograders cannot handle partial credit, equivalent solutions, or syntactic variation. This paper evaluates whether four frontier Large Language Models (GPT, Claude Opus, Gemini, and GLM) can approximate expert judgment when grading short Linux/bash command responses. The study adopts a four-level cognitive taxonomy that combines cognitive complexity and operational impact, ranging from information retrieval (L1) and basic file manipulation (L2) to structural operations (L3) and advanced system management (L4). The models were tested with two prompt variants, a minimal baseline and a rubric-enhanced version, on 1200 real responses from second-year Computer Engineering students independently graded by three expert instructors. Gemini~3.0 Pro with rubric-guided prompting achieved the highest human-AI agreement (ICC(3,1) = 0.888, MAE = 0.10, Bland-Altman bias = $-0.014$). Agreement declined consistently as taxonomy level increased, with the largest discrepancies at higher levels. Across all models, rubric quality had a larger effect than provider choice, with structured prompts consistently improving agreement. These results show that question complexity is a reliable predictor of the difficulty LLMs face in grading accurately, and they establish a principled, taxonomy-based framework for determining which questions are suitable for AI-assisted grading and which require human review, while also providing a transferable evaluation protocol and prompt templates.

\end{abstract}




\end{frontmatter}

\section{Introduction}

Computer science education continues to face strong scalability pressures due to the growing demand for these degree programs~\citep{eurostat_ict_education_overview}. The high number of students increases the volume of coursework and tests, making it increasingly difficult to maintain high-quality assessment and feedback within a reasonable turnaround time~\citep{Bernik_2025,Siek_2026}. Autograding systems and reference outputs address functional correctness efficiently, but they often provide limited support for partial credit, alternative correct solutions, and explanations that help students improve. As a result, there is growing interest in hybrid assessment pipelines that preserve test based checks while using Large Language Models (LLMs) to generate feedback and to support rubric guided grading in a way that approximates instructor judgment~\citep{Pereira_2025}.

This challenge is particularly acute in courses that evaluate command line proficiency because student answers are typically short text responses rather than complete software snippets. Small syntactic choices can change semantics, and many tasks admit multiple equivalent commands. In addition, partially correct answers are common: a response may be conceptually appropriate but include missing flags, incorrect quoting, or mistakes in ordering within a pipeline. These characteristics make command line assessment difficult to capture with strict matching rules, while also making manual grading time consuming and sensitive to grader interpretation.

LLMs offer a promising mechanism for handling open form technical responses, because they can interpret natural language and code-like text, reason over constraints, and provide justifications. However, the central question is not whether an LLM can produce plausible feedback, but whether its scores are reliable and aligned with human evaluation under a specified rubric and grading scale. Prior studies in programming and constructed response grading show that alignment depends on prompt design and rubric specificity, and that performance can vary across models and tasks~\citep{Pathak_2025,Bolgova_2025,Qiu_2026}. However, most evidence focuses on full programming assignments, whereas short-form command-line responses introduce distinct challenges: multiple equivalent solutions, high sensitivity to syntactic variations, and frequent partially-correct answers.

This paper addresses a domain specific evaluation of LLM assisted grading for bash command exams, a task that combines the ambiguity of constructed response assessment with the syntactic precision of code evaluation. Much of the existing literature focuses on grading full programming assignments, where responses are longer code artifacts and correctness can often be checked via compilation and unit tests~\citep{Raihan_2025}. However, competence with the bash shell is equally central in operating systems and systems administration curricula, and the assessment of short bash command responses is comparatively less studied. These responses pose distinct challenges, including multiple equivalent solutions and high sensitivity to small syntactic variations, which make them difficult to evaluate with either strict matching rules or unconstrained LLM scoring. To fill this gap, four state of the art LLMs are compared against a three-evaluator human baseline on a real examination dataset of 1200 responses of basic Linux/bash system administration commands, and prompt design choices are systematically examined to determine how they modulate agreement. 

The remainder of this work is organized as follows. Section~\ref{sec:related} reviews prior work on automated assessment and human-AI agreement in educational settings. Section~\ref{sec:methodology} describes the dataset, the three-evaluator human grading protocol, the evaluated LLMs, and the prompting variants used for AI assessment. Section~\ref{sec:results} presents the agreement and error analyses comparing human and LLM evaluations, and Section~\ref{sec:discussion} discusses the implications, limitations, and practical considerations for adoption. Finally, Section~\ref{sec:conclusion} summarizes the main contributions and outlines directions for future research.

\section{Related Work}
\label{sec:related}

The following subsections review LLM-based grading approaches and the methodological foundations for quantifying agreement between humans and AI.

\subsection{LLM-Based Assessment in Education}

Automated assessment has become central to scaling computer science education. While traditional autograding pipelines efficiently check functional correctness via unit tests, they struggle with multi-criteria evaluation and pedagogically useful feedback. Recent work explores hybrid approaches using LLMs to approximate richer instructor feedback~\citep{Wang2024,Pereira_2025}.

In programming assignments, several frameworks arose that treat LLMs as structured graders constrained by rubrics. CodEv combines Chain-of-Thought (CoT) prompting with LLM ensembles and agreement tests~\citep{Tseng_2024}, while other approaches fine-tune LLMs to improve grading consistency~\citep{Yousef_2025}. Classroom deployments show mixed results: Mark My Works found no significant correlation between AI and human scores in a 191-student pilot, with AI grading more conservatively~\citep{Qiu_2026}, whereas multi-course evaluations using structured rubrics report high AI--instructor correlation~\citep{Bernik_2025}. 

Prompting strategy significantly affects grading quality. CoT prompting improves interpretability and alignment with expert evaluation~\citep{Akyash_2025}, while question-specific rubrics enhance agreement, with studies introducing leniency metrics to quantify systematic strictness differences~\citep{Pathak_2025}. Work comparing LLM judgments to human graders confirms that prompt engineering and careful feature selection position LLMs as complementary tools rather than replacements~\citep{Grandel_2024}. Benchmarking studies reveal substantial model-dependent differences in accuracy~\citep{Mohamed_2025}, and feedback generation, while effective, may include hallucinated issues~\citep{Jacobs2024}.

Beyond programming, LLMs have been evaluated for short-answer grading~\citep{Wangwiwattana_2023} and essay scoring~\citep{Pack_2024,Gaggioli2025}. Results vary by domain, as an example: medical education comparisons show mixed effects of rubric provision on agreement metrics, while essay scoring exhibits temporal fluctuations and weak cross-replication stability~\citep{Bolgova_2025}. Fine-tuning via parameter-efficient methods (LoRA, QLoRA) improves scoring accuracy and feedback similarity to experts~\citep{Ashiya_Katuka_2024}, and specialized datasets such as EngSAF demonstrate successful real-world deployment~\citep{Aggarwal_2025}.

Overall, LLM-based assessment can range from high alignment to substantial divergence depending on rubric quality, prompt design, and domain. The empirical evidence concentrates on programming assignments and essays, leaving systematic evaluation of command-line assessment unexplored.

\subsection{Quantifying Agreement Between Humans and AI}

Human--AI agreement studies frame grading as a measurement problem, quantifying whether model scores are interchangeable with human judgments~\citep{Wang_2025,Pecuchova_2025,Emirtekin_2026}. The choice of metric must match the data type and intended decision.

For categorical outcomes (pass/fail, grade bands), chance-corrected measures such as Cohen's $\kappa$ distinguish substantive concordance from chance agreement. For ordinal scales, Weighted Kappa (WK) credits near-matches while penalizing large discrepancies~\citep{Emirtekin_2026}. Multi-rater settings employ Fleiss' $\kappa$ or Krippendorff's $\alpha$ alongside error profiles~\citep{Pecuchova_2025}. For continuous scores, Pearson correlation captures linear association between AI and human scores, while Spearman correlation evaluates whether both evaluators preserve the same rank ordering even when the relationship is not strictly linear~\citep{Freedman_2007,Spearman_1904}. Also, Intraclass Correlation Coefficient (ICC) assesses both absolute agreement and consistency in relative ordering~\citep{Shrout_1979}, with different model specifications (ICC(2,1), ICC(3,1)) addressing whether raters are random or fixed.

Beyond agreement coefficients, psychometric frameworks using generalizability theory and many-facet Rasch modeling separate variance sources and analyze rater severity~\citep{Wang_2025}, while related work evaluates AI-supported rating for replicability~\citep{Pillet_2025}. Agreement metrics should also be interpreted alongside fairness considerations, as discrepancies may manifest systematically across subgroups~\citep{Arshad_2025}, and stakeholder acceptance, since students may prefer human grading even when AI scores are accurate~\citep{Thomas_2026}. For human vs AI analysis, Bland--Altman is especially informative because, unlike Fleiss' $\kappa$ , ICC, or simple correlations, it directly tests whether AI and human scores are practically interchangeable by exposing systematic bias and the limits of agreement across the score range~\citep{Bland_1986}.

\section{Materials and Methodology}\label{sec:methodology}

\begin{figure}[!h]
    \centering
    \includegraphics[width=\textwidth]{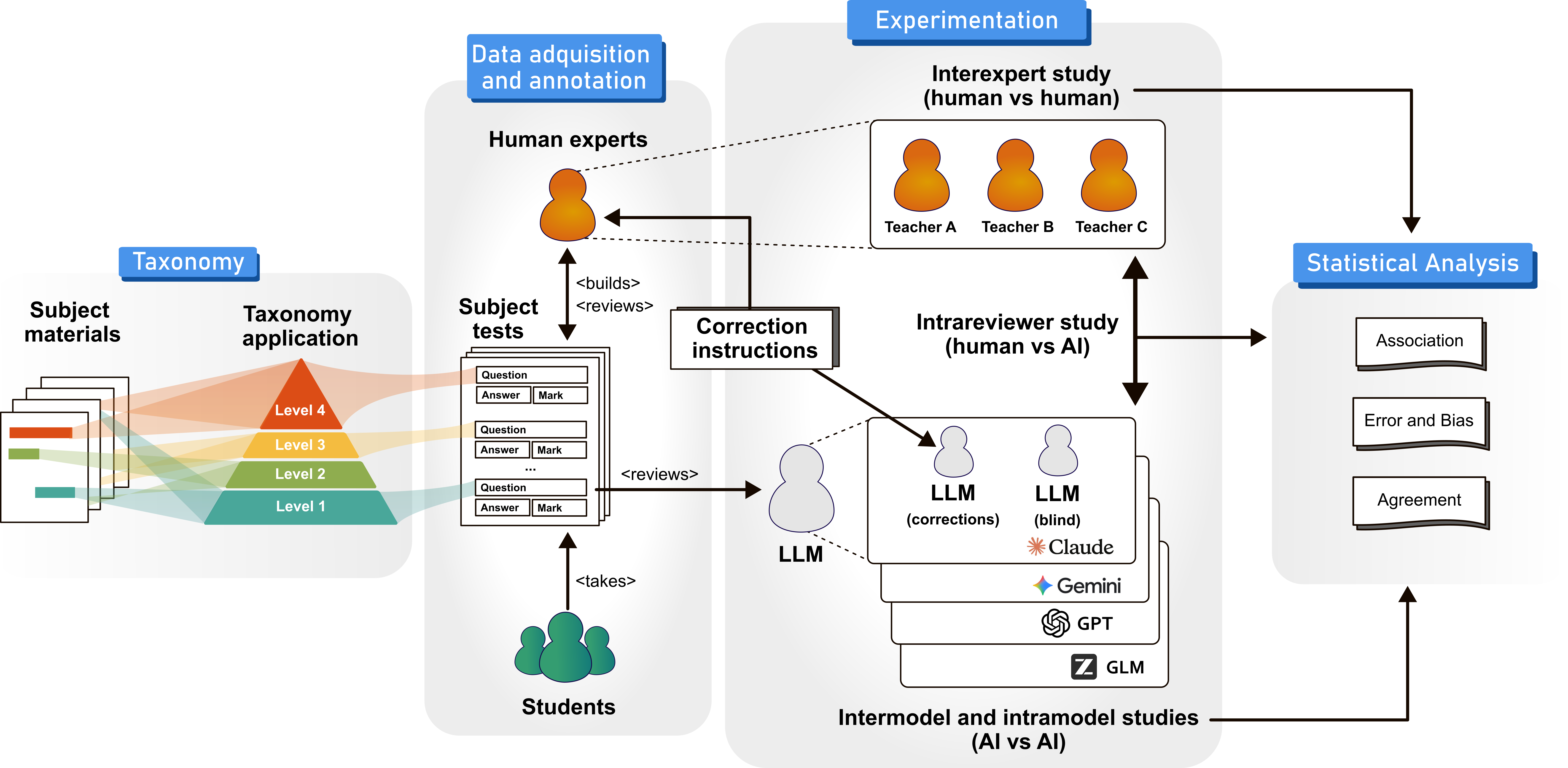}
    \caption{Research methodology.}
    \label{fig:proposal}
\end{figure}

The methodology employed in this research follows a systematic approach designed to compare automated AI-based assessment with human evaluation in the context of Linux/bash command exams. The process, as shown in Figure~\ref{fig:proposal}, begins with the classification of all examination questions according to a four-level taxonomy (Section~\ref{subsec:taxonomy}) that combines Bloom's Revised Taxonomy with an operational-impact dimension specific to system administration tasks. Since the exam questions are drawn from the subject materials, each question is tagged with its corresponding taxonomy level, enabling granular analysis of across different levels of command complexity. Then, after the students have undergone the test, the collection of student responses to examination questions covering the various bash commands and operations taught throughout the course are collected.

Once student responses are gathered, the first phase of evaluation is conducted by human evaluators. Three experienced instructors independently grade each student response following a predefined rubric that considers factors such as syntactic correctness, functional adequacy, efficiency, and conceptual understanding. The human evaluators assign numerical scores to each response, providing a baseline for comparison. This human evaluation approach with three evaluators enables the calculation of inter-rater reliability metrics, which serves to establish the consistency and objectivity of the human grading process before comparing against AI-based systems. The full description of the dataset, obtained through this process, is covered in Section~\ref{subsec:dataset}. These responses represent a diverse range of student performance levels and command categories, ensuring comprehensive coverage of the assessment spectrum.

Following the completion of human assessment, the same student responses are submitted to multiple LLMs for automated evaluation. Several state of the art models, including Claude Opus~4.6 (Anthropic), GPT~5.2 (OpenAI), Gemini~3.0 Pro(Google), and GLM~5 (Zhipu AI), are employed as automated evaluators. Each model, described in Section~\ref{subsubsec:llm_models}, receives identical prompting instructions and contextual information about the course objectives. This different input variants are deeply explained in Section~\ref{subsubsec:eval_process}. The models independently generate a score and justification for each student response, mirroring the evaluation process performed by human experts. This parallel evaluation structure allows for direct comparison between human and AI-generated assessments.

The final phase involves comprehensive statistical comparison of the evaluation results. Multiple statistical metrics are computed to assess agreement, correlation, and systematic bias between human evaluators and LLM models (Section~\ref{subsec:human_evaluation}). Inter-rater reliability measures such as Intraclass Correlation Coefficient (ICC), Weighted Kappa (WK), Pearson and Spearman correlation coefficients, Mean Absolute Error (MAE), and Bland-Altman analysis are applied to quantify the level of concordance. These comparisons are performed both globally across all responses and stratified by taxonomic level, enabling identification of specific command categories where AI evaluation may diverge from human judgment. The results of these comparisons provide empirical evidence regarding the viability and limitations of AI-powered automated assessment in technical computing education contexts. This is closely analyzed in Section~\ref{sec:results}.

This methodology relies on classifying each examination question by its level of cognitive and operational demand, so that agreement between human and automated evaluators can be analysed not only in aggregate but also as a function of question complexity. This classification is provided by the four-level taxonomy described in the next subsection. 

\subsection{Taxonomy}\label{subsec:taxonomy}
\begin{figure}[!h]
    \centering
    \includegraphics[width=\textwidth]{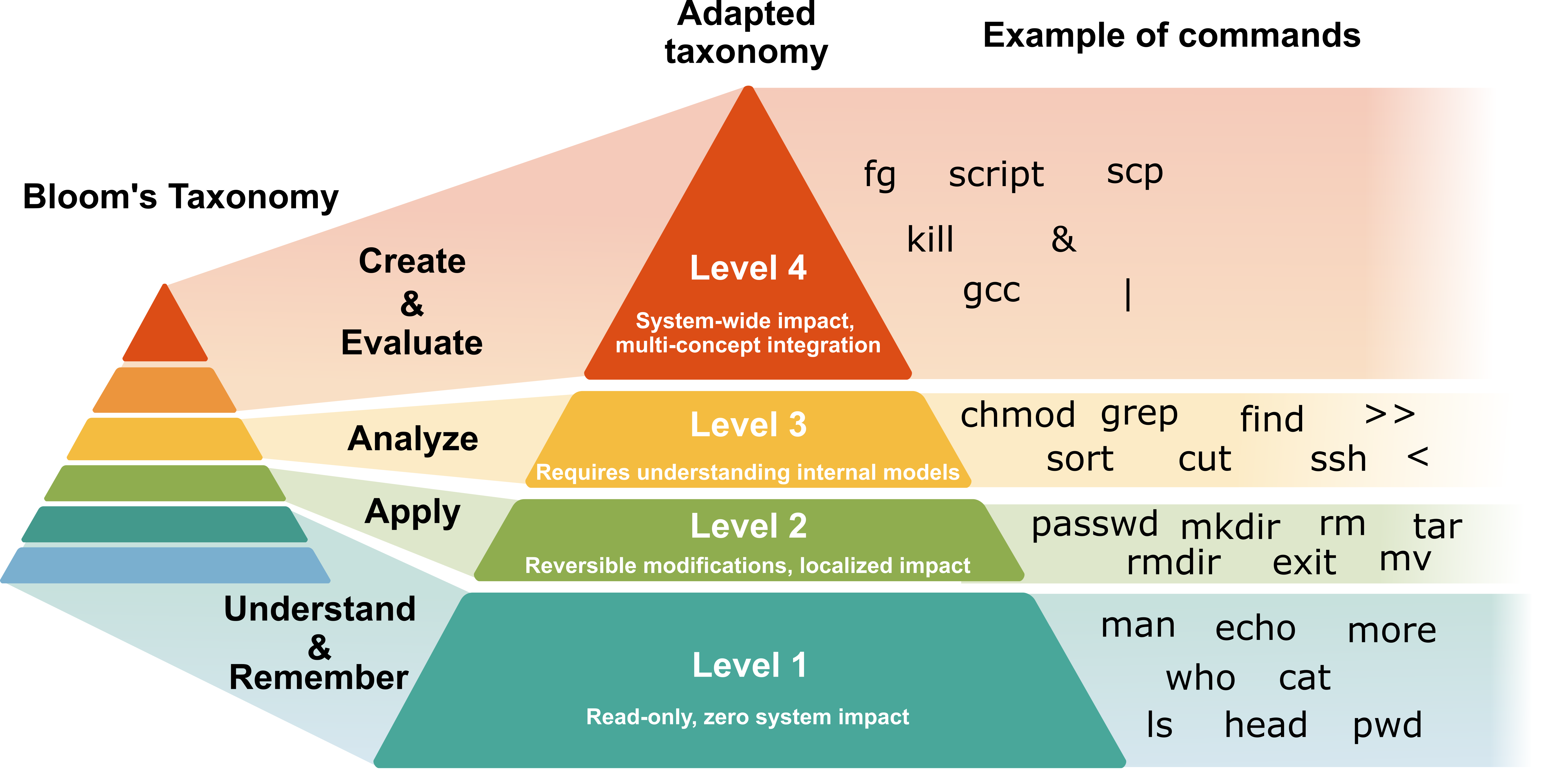}
    \caption{Used taxonomy to classify commands depending on its cognitive complexity and operational impact.}
    \label{fig:bloom_adapted}
\end{figure}

To enable meaningful analysis of both human and automated evaluation, we use a four-level taxonomy that classifies the commands covered in the course contents based on the cognitive processes and conceptual integration required for their proper use. This taxonomy, known as CogTax, represented in Figure~\ref{fig:bloom_adapted} and proposed previously in ~\citep{AlonsoCarracedo2026}, ranges from basic information retrieval operations to complex tasks involving the integration of multiple concepts and inter-process communication. By categorizing commands according to this framework, it is possible to systematically examine whether evaluators, both human and AI, apply consistent criteria across different levels of complexity, and identify patterns in grading discrepancies that may correlate with the cognitive demands of specific command categories.

CogTax comprises four different categories designed following the principles of Bloom's Taxonomy. Since its inception in 1956 and subsequent revision in 2001, Bloom's Taxonomy has been widely adopted across various disciplines, including computer science, to guide curriculum design and evaluation ~\citep{Philip_Machanick_2000, JiChuan_Quan_2017}. 
Bloom’s hierarchical cognitive levels (remembering, understanding, applying, analyzing, evaluating, and creating) and its computing-specific enhancement~\citep{acm2023}, serve as lenses to evaluate instructional methods and learning outcomes in basic bash instructions~\citep{Gregory_Kirk_Johnson_2012, Mohsen_Dorodchi_2017}.

This four-level taxonomy integrates two dimensions. First, \textit{cognitive complexity (C)}, that is, the mental operations required to use commands effectively (derived from Bloom's Revised Taxonomy) and second, \textit{operational impact (O)}, that is, the degree to which commands modify system state and the reversibility of those modifications. The integration of these two dimensions is defined by Equation~\eqref{eq:taxonomy}. This formula ensures monotone coverage: a command at level $L$ requires at least level-$L$ understanding \textit{or} produces at least level-$L$ effects, but not necessarily both. The consequence is that instructors and students must address whichever dimension is higher as conceptual mastery alone is insufficient if operational awareness is absent, and operational awareness alone is insufficient without conceptual understanding.

\begin{equation}
    L = \max(C, O)
    \label{eq:taxonomy}
\end{equation}

The four levels of the taxonomy are defined below. Each level is characterized by the cognitive operations it demands, the operational impact it produces on the system, and the pedagogical role it plays in a Linux/bash curriculum. The levels are ordered by increasing complexity, so that each successive level requires either a higher cognitive load, a greater potential for irreversible system change, or both.

\begin{description}

\item{\textbf{Level 1: Information Query / Observation.}}
L1 commands retrieve or display information without modifying system state. They are read-only by design: no file is created, no process altered, no permission changed. Commands in this level include \texttt{ls}, \texttt{cat}, \texttt{pwd}, \texttt{echo}, or \texttt{cal}. Cognitively, L1 commands require vocabulary recognition and basic shell syntax. Operationally, they are reversible by definition. L1 commands form the safe exploration zone of any computing course. Students can interact with the system freely, build mental models of directory structure and file content, and develop pattern recognition without risk of irreversible change. Assessments at L1 test whether students can read system state and interpret command output.

\item{\textbf{Level 2: Basic Modifications.}}
L2 commands introduce well-defined side effects: creating, copying, moving, or removing files; setting variables; applying simple redirections. Commands in this level include \texttt{mkdir}, \texttt{cp}, \texttt{mv}, or \texttt{passwd}. Cognitive complexity is moderate (students must model preconditions and postconditions); operational impact is bounded and mostly reversible. L2 commands introduce state management and marks the beginning of functional thinking about transformations. Students learn to reason about what a command requires as input and what it produces as output. Assessments at L2 test whether students can predict the postcondition of a command from its precondition.

\item{\textbf{Level 3: Structural Understanding.}}
L3 commands involve multiple operators, conditional logic, pipelines, or permission structures. Commands in this level include \texttt{chmod}, \texttt{find} or simple pipelines (\texttt{command1 | command2}). Cognitive complexity is high (composing multiple primitive concepts); operational impact may be significant. L3 commands represent the productive challenge zone for intermediate learners. They require both conceptual integration and operational awareness which are essential for understanding how combined effects propagate through the system. 

\item{\textbf{Level 4: Advanced System Management.}}
L4 commands are deeply nested, multi-concept operations with high cognitive and operational complexity. Commands in this level include \texttt{kill}, \texttt{scp} or background process management (\texttt{\&}). These commands require simultaneous reasoning across multiple layers of abstraction. L4 commands define the expert horizon, appropriate for setting the foundation for advanced courses. Assessment at L4 requires the ability to decompose complex compound operations, reason about interaction effects, and trace execution across concurrent processes.
\end{description}

The full list of which commands belong to each level can be found in ~\citep{AlonsoCarracedo2026}.

\subsection{Dataset}\label{subsec:dataset}

The experimental dataset consists of 1200 exam responses collected from second-year undergraduate students enrolled in a Computer Engineering program, specifically an Operating Systems course. The examination was administered as a closed-book, 90-minute assessment designed to evaluate student proficiency in Linux bash command-line operations. This assessment represents a midterm evaluation conducted after four weeks of intensive instruction, during which students received structured classroom teaching complemented by independent practice exercises which covered the four taxonomic levels. 

The examination was conducted in a controlled server environment where all students simultaneously accessed a shared Linux system. It comprised 16 independent exercises, that were classified according to the four-level cognitive taxonomy. Each student created their response document directly on the server during the 90-minute examination period, ensuring identical testing conditions and preventing external resource access. Upon completion, all student response documents were collected from the server and subsequently imported into a custom-developed web-based evaluation management platform. This platform, specifically designed for educational assessment research, provides a comprehensive examination management system that facilitates question bank management, response collection, multi-evaluator correction workflows, and data export capabilities. The platform supports systematic tagging of examination questions according to taxonomic classifications, enabling precise organization and subsequent analysis of questions based on cognitive complexity levels.

Once imported into the platform, each response was automatically assigned a unique identifier while maintaining student anonymization capabilities for blind evaluation purposes.  

Then, the human evaluation process was conducted by three experienced human instructors with backgrounds in Linux system administration and command-line instruction. All three human evaluators independently assessed all student responses following a comprehensive rubric developed and refined over multiple academic terms based on common student errors and performance patterns observed in previous years.

The assessment rubric operates at two complementary levels. First, it establishes general criteria applicable across all examination questions, addressing fundamental aspects such as command structure, syntax correctness, appropriate use of file paths, proper sequencing of pipeline operations, and overall readability of responses. These general considerations ensure consistent evaluation of basic technical competencies regardless of specific question content. Second, the rubric provides question-specific scoring guidelines tailored to each examination item, with point values ranging from 0.25 to 1.25 depending on the complexity and scope of the required response.

For each examination question, the rubric documents one or more acceptable correct solutions, accounting for the diversity of valid approaches available in bash command-line operations. Additionally, the rubric catalogs frequently observed student errors specific to each question, along with their corresponding point deductions. 

All three evaluators performed their assessments independently using the platform's blind evaluation interface, which conceals evaluator identities and prevents access to the other evaluators' scores during the grading process. This independent evaluation protocol minimizes potential bias and enables subsequent calculation of inter-rater reliability metrics to establish the consistency and objectivity of human assessment before comparison with automated AI-based evaluation. 

The final dataset therefore contains, for each student response: the question text, the student's free-text answer, the maximum achievable score, the per-question rubric with its partial-credit deductions, one or more accepted correct solutions, and the independent grades and comments from all three human evaluators. When submitting responses to the LLMs, however, only a subset of this information is provided, and the exact subset depends on the prompt variant further explained in Section~\ref{subsubsec:eval_process}.

The same set of responses was then submitted to the four LLMs, and their scores were compared against the human consensus using the same statistical battery applied to the human evaluators. The metrics used for both comparisons are described in the following subsection.

\subsection{Evaluation metrics}\label{subsec:human_evaluation}

As reviewed in Section~\ref{sec:related}, no single metric is sufficient to fully characterize inter-rater agreement. These metrics can be divided into the categories association, absolute agreement, error and bias, and ordinal agreement, each capturing a distinct facet of concordance. The following subsections describe the specific metrics selected for each category.

\subsubsection{Association}

Association measures assess whether the scores from one evaluator correspond to higher scores from another, without requiring that the evaluators assign identical values. Strong association is a necessary but not sufficient condition for agreement, because two evaluators can be perfectly correlated yet differ by a constant offset.

Pearson's product-moment correlation coefficient ($r$) was calculated to measure the strength and direction of the linear relationship between evaluators' scores ~\citep{Freedman_2007}. This metric assesses whether evaluators maintain proportional scoring patterns across the range of student performance levels. Values range from -1 to +1, where values closer to +1 indicate strong positive linear association.

Spearman's rank correlation coefficient ($\rho$) was employed to evaluate the monotonic relationship between rankings of the evaluators. Unlike Pearson's correlation, Spearman's metric is robust to non-linear relationships and outliers, as it operates on rank-transformed data rather than raw scores. This provides complementary insight into whether evaluators maintain consistent relative ordering of student performance even when absolute score assignments may differ.

\subsubsection{Absolute Agreement}

Absolute agreement measures go beyond association by requiring that evaluators assign not merely correlated but interchangeable scores. A high value indicates that any evaluator drawn from the same population would produce essentially the same grade for a given student.

The Intraclass Correlation Coefficient with a two-way random-effects model for single measures, ICC(2,1), was computed to assess absolute agreement between human evaluators. This metric quantifies the proportion of total variance attributable to true differences between students rather than inconsistencies between evaluators. ICC values are interpreted following established guidelines: values below 0.50 indicate poor reliability, 0.50-0.75 moderate reliability, 0.75-0.90 good reliability, and above 0.90 excellent reliability. The ICC(2,1) model is particularly appropriate for this study as it treats both students and evaluators as random samples from larger populations, allowing generalization beyond the specific evaluators employed.

In Section~\ref{sec:discussion}, when comparing LLMs against the human consensus ICC(3,1) is used. It focuses on consistency rather than absolute agreement. Unlike the human evaluators, the four models are not interchangeable samples from a larger population; they are the specific systems under evaluation. It is therefore only necessary to check whether each model ranks and spaces students in the same way the human evaluators do, even if its raw scores are systematically higher or lower. A constant offset (for instance, a model that grades every student 0.3 points below the human consensus) is a straightforward calibration issue that can be corrected by a simple additive adjustment; what cannot be easily corrected is a model that misjudges the relative ordering of students. ICC(3,1) targets precisely that distinction, while the magnitude of any systematic shift is already captured by error and bias metrics.

\subsubsection{Error and Bias}

Error and bias measures complement the preceding coefficients by quantifying how far apart the scores of the evaluators actually are and whether the differences are systematic. While the previous metrics produce dimensionless indices, these measures express disagreement in the original grade units, offering a directly interpretable picture of practical scoring discrepancies.

Mean Absolute Error (MAE) quantifies the average magnitude of differences between scores without regard to direction. This metric provides an intuitive measure of typical scoring discrepancy expressed in the same units as the original grades. Lower MAE values indicate greater consistency in absolute score assignments.

Bland-Altman analysis was performed to assess systematic bias and limits of agreement between evaluators ~\citep{Bland_1986}. With three evaluators, Bland-Altman comparisons are performed pairwise and summarized across evaluator pairs. This approach plots the difference between paired evaluator scores against their mean, enabling visualization of whether disagreements are consistent across the scoring range or concentrated at particular performance levels. The analysis computes the mean difference (systematic bias) and 95\% limits of agreement (mean difference $\pm$ 1.96 × standard deviation of differences), providing bounds within which 95\% of evaluator disagreements are expected to fall. This statistical approach is particularly effective at revealing whether one evaluator systematically assigns higher or lower scores than another, and whether agreement deteriorates at extreme performance levels.

\subsubsection{Ordinal Agreement}

Ordinal agreement measures account for the fact that grading categories have a natural order. A one-category discrepancy represents a less severe disagreement than a multi-category one. These metrics credit near-matches while penalizing large deviations more heavily, making them especially suited to educational scales where partial credit is meaningful.

Weighted Kappa~\citep{Cohen_1968} extends classic Cohen's Kappa~\citep{Cohen_1960} by incorporating the magnitude of disagreement. Classic Cohen's Kappa is designed for exactly two raters, when three or more evaluators are involved, it cannot be applied directly and pairwise computation becomes necessary.

Rather than treating all disagreements equally, Weighted Kappa assigns partial credit when categorizations are close but not identical, and pairwise scores are subsequently summarized across all evaluator pairs. Quadratic weighting was employed, which increasingly penalizes larger discrepancies. This metric is particularly valuable in educational assessment contexts because it treats nearby grades as more similar than distant ones, assigning progressively greater penalties as the discrepancy between ratings increases.

Together, the four metric categories described above form a complementary battery capable of detecting the different ways in which an automated evaluator may diverge from expert human judgment. This battery is applied first to quantify consistency among the three human evaluators, establishing a reliability ceiling, and then to compare each LLM against the human consensus. The LLM used are introduced in the following section.

\subsection{LLM Models}
\label{subsubsec:llm_models}

Four LLMs were evaluated, including GPT~5.2, Claude Opus~4.6, Gemini~3.0 Pro, and GLM~5. These models were selected to represent a diverse range of provider ecosystems, ensuring that the evaluation captures a broad spectrum of contemporary model capabilities. The goal is not to claim that any single model is universally best, but to evaluate representative frontier systems that are widely deployed and that reflect different design and training philosophies. By evaluating multiple models under the same conditions, it is possible to assess whether observed effects are consistent across architectures or specific to particular model families.
\begin{description}
    \item[GPT~5.2 (OpenAI):] GPT class models are among the most commonly deployed general purpose LLMs in education oriented tools and developer workflows. GPT~5.2~\citep{OpenAI_GPT52_2026} is included as a representative high capability model for instruction following and code related reasoning.
    
    \item[Claude Opus~4.6 (Anthropic):] Claude's Opus family is a flagship line that is frequently used for long form reasoning and careful natural language explanations. Claude Opus~4.6 \citep{Anthropic_ClaudeModels_2026} is included to test whether a model that is often perceived as strong in deliberative writing and nuanced explanation exhibits different grading tendencies under the same rubric constrained prompts. 
    
    \item [Gemini~3.0 Pro (Google):]Gemini models are widely deployed across productivity and developer ecosystems and are commonly used for code-related assistance. Gemini~3.0 Pro~\citep{Google_Gemini3ProPreview_2026} is a frontier model from a separate provider and model family, to assess whether agreement with expert grading is robust across ecosystems rather than specific to a single vendor. This is particularly relevant for educational stakeholders who may have institutional constraints that favor certain cloud providers.
    
    \item [GLM~5 (Zhipu AI):] GLM models represent an additional modern LLM lineage with different training data mixtures and alignment approaches compared to the previous models. GLM~5~\citep{ZhipuAI_ModelOverview_2026} adds provider diversity and allows this study to evaluate whether rubric guided grading and prompt design effects generalize beyond the most commonly studied provider models. 
\end{description}

\subsection{Evaluation Process and Prompt Ablation Study}
\label{subsubsec:eval_process}

To isolate the contribution of rubric provision to grading quality, each of the four models was evaluated under two prompt variants applied to the same set of student responses. The two variants differ in the amount of structured guidance provided: the first (Variant 1 - No Rubric) supplies the model with the minimum context needed to grade, while the second (Variant 2 - With Rubric) additionally provides the full rubric and a reference correct answer. All models used their default configurations to ensure reproducibility and eliminate confounding variables related to parameter tuning. This controlled contrast isolates the effect of rubric provision on grading quality, making it possible to quantify how much of the observed human--AI agreement is attributable to model capability alone versus the explicitness of the evaluation criteria.

\subsubsection{Variant 1 - No Rubric}

This prompt variant represents the minimal configuration baseline for the evaluation framework. It establishes a foundation for comparison by providing the model with only the essential elements needed to perform the grading task: the evaluator's role, the course context materials, the exam question, the student's response and the maximum possible grade of each question.

The baseline prompt deliberately omits a detailed grading rubric with explicit scoring criteria. This minimal approach enables assessment of the model's intrinsic ability to evaluate bash command responses based solely on its pre-trained knowledge and general understanding of educational assessment practices, augmented only by the specific course materials and max grades.

By excluding structured grading guidelines and exemplar corrections, this baseline configuration tests whether models can perform competent evaluation through inference alone, mimicking a scenario where an instructor evaluates without explicit rubrics thus relying on expertise and general pedagogical principles. The performance metrics obtained from this variant serve as a reference point for measuring the marginal improvements achieved by adding structured guidance (Variant 2).

All models evaluated in this study use their default configurations (temperature, top-p, etc.) to ensure reproducibility and eliminate confounding variables related to parameter tuning. The following listing (Listing \ref{prompt1}) shows the general structure of the prompt used for this variant.

\begin{lstlisting}[basicstyle=\scriptsize\ttfamily, caption={Variant 1 - No Rubric prompt structure},label={prompt1}] 

You are a rigorous and concise examiner of operating systems exams.
Evaluate answers precisely and technically, penalizing syntax errors
and omissions. ALWAYS respond in this exact format:

Grade: <number>
Justification: <text>

Additional instructions:
Assign partial credit only if the answer demonstrates understanding of the concept despite
minor errors. 
The grade must be a number between 0 and the maximum grade indicated in 0.05 steps.

{retrieved_course_materials}

Question {letter}:
{question_text}

Student answer:
{student_answer}

Maximum grade for this question: {max_grade}

Correct answer: {correct_answer}
\end{lstlisting}

\subsubsection{Variant 2 - With Rubric}

This prompt variant offers the LLM models the same information that the human evaluator have, offering a direct comparison between them. It introduces explicit evaluation criteria through the same grading rubric used by human evaluators, representing the first systematic enhancement over the baseline configuration. The addition of a structured rubric addresses one of the fundamental challenges in automated educational assessment: ensuring consistent, transparent, and criterial evaluation across all responses. 

The rubric implemented in this variant provides general evaluation considerations such as completeness, command syntax, and correct path definition, while also incorporating question-specific aspects such as alternative valid solutions and common mistake penalties. This information serves a dual purpose: it helps the model recognize frequent error patterns and provides consistency in how these errors impact the final grade. Additionally, the prompt includes explicit instructions for evaluating alternative solutions, acknowledging that many bash commands have multiple valid implementations.

The rubric-enhanced configuration tests the hypothesis that structured guidance improves LLM grading accuracy by reducing ambiguity in evaluation criteria and providing explicit standards for quality assessment. By comparing performance between Variant 1 and Variant 2, it is possible to measure the isolated effect of criterial structure on model agreement with human expert evaluations. This comparison is particularly relevant for educational institutions considering AI-assisted grading, as it quantifies the value of investing effort in rubric development.

The prompt is extended in this variant to include penalty specific considerations and alternative answers, as shown in listing \ref{prompt2}.

\begin{lstlisting}[basicstyle=\scriptsize\ttfamily, , caption={Variant 2 - With Rubric prompt structure},label={prompt2}]

You are a rigorous and concise examiner of operating systems exams.
Evaluate answers precisely and technically, penalizing syntax errors
and omissions. ALWAYS respond in this exact format:
Grade: <number>
Justification: <text>

Additional instructions:
Compare the answer with the correct solution. Assign partial credit
only if the answer demonstrates understanding of the concept despite
minor errors. The grade must be a number between 0 and the maximum
grade indicated in 0.05 steps.

{retrieved_course_materials}

Question {letter}:
{question_text}

Student answer:
{student_answer}

Maximum grade for this question: {max_grade}

Rubric: {rubric}

Correct answer: {correct_answer}
\end{lstlisting}

\section{Results}
\label{sec:results}

The preceding sections have established a comprehensive experimental framework: a four-level cognitive taxonomy grounded in CogTax and operationally extended to capture bash cognitive complexity and operational impact on the system; a real-world dataset of 1200 student exam responses spanning the full performance spectrum, collected under controlled conditions and independently graded by three experienced instructors (Section~\ref{subsec:dataset}); four state-of-the-art LLMs evaluated under two prompt variants of increasing rubric specificity (Section~\ref{subsec:human_evaluation}); and a battery of complementary statistical metrics, ranging from correlation and ICC to WK, MAE, and Bland--Altman analysis, designed to quantify agreement, consistency, and systematic bias at multiple levels of granularity. Building on this infrastructure, the present section reports the empirical results of the evaluation study. We first characterize the human grading baseline, reporting descriptive statistics and inter-rater reliability metrics that establish the reference standard against which automated assessments are compared (Section~\ref{subsec:human_consensus}). We then present the grades and score distributions produced by each LLM under both prompt variants (Section~\ref{subsec:llm_results}). A detailed analysis and interpretation of these results, including their implications for AI-assisted assessment practice, is provided in Section~\ref{sec:discussion}.

The present section reports the empirical results of the statistical metrics computed. The human grading baseline is first characterized, reporting descriptive statistics and inter-rater reliability metrics that establish the reference standard against which automated assessments are compared (Section~\ref{subsec:human_consensus}). We then present the grades and score distributions produced by each LLM under both prompt variants (Section~\ref{subsec:llm_results}). A detailed analysis and interpretation of these results, including their implications for AI-assisted assessment practice, is provided in Section~\ref{sec:discussion}.

\subsection{Human Consensus}\label{subsec:human_consensus}

This section characterizes the human grading baseline against which all automated assessments are subsequently compared. The frequency distribution of item-level scores is first examined to reveal how expert evaluators allocate marks across the response quality spectrum (Section~\ref{subsubsec:score_dist_humans}). Performance is then stratified by taxonomy level to show how scoring varies with cognitive complexity (Section~\ref{subsubsec:taxonomy_humans}). Finally, aggregate inter-rater reliability metrics are reported to quantify the consistency among the three expert evaluators and establish the psychometric ceiling that automated methods must approach (Section~\ref{subsubsec:stats_humans}).

To facilitate interpretation of the grade distributions reported in this section, the standard Spanish university grading scale is adopted, which maps numerical scores on a 0--10 continuum to four categorical bands: \textit{Fail} (\textit{Suspenso}, $[0, 5)$), \textit{Pass} (\textit{Aprobado}, $[5, 7)$), \textit{Merit} (\textit{Notable}, $[7, 9)$), and \textit{Distinction} (\textit{Sobresaliente}, $[9, 10]$). For readers more familiar with letter-grade systems, these bands correspond approximately to F (Fail), C--D (Pass), B (Merit), and A (Distinction) in the US grading scale, or to Fail, Third/Lower Second, Upper Second, and First in the UK classification system. This categorization is used throughout the descriptive statistics tables (e.g., Table~\ref{tab:human_descriptive}) to summarize the proportion of student final grades falling into each qualitative band.

\begin{table}[h]
\centering
\caption{Descriptive statistics of human evaluator final grades.}
\label{tab:human_descriptive}
\small
\begin{tabular}{llccccccccc}
\toprule
\textbf{Eval.} & \textbf{Mean} & \textbf{Median} & \textbf{Min/Max} & \textbf{Zeros} & \textbf{Perfect} & \textbf{Fail} & \textbf{Pass} & \textbf{Merit} & \textbf{Dist.} \\
\midrule
A & 5.33 & 5.50 & 0.0\,/\,10 & 356 & 550 & 43.7\% & 25.3\% & 25.3\% & 5.7\% \\
B & 5.36 & 5.52 & 0.0\,/\,10 & 357 & 550 & 44.8\% & 25.3\% & 24.1\% & 5.7\% \\
C & 5.54 & 5.94 & 0.0\,/\,10 & 338 & 571 & 39.1\% & 25.3\% & 31.0\% & 4.6\%\\
\midrule
Mean & 5.41 & 5.70 & 0.0\,/\,10 & 350 & 557 & 42.5\% & 26.4\% & 25.3\% & 5.3\% \\
\bottomrule
\end{tabular}
\end{table}

Table~\ref{tab:human_descriptive} summarizes the descriptive statistics of the consensus grades assigned by the human evaluators across the 1,200 gradable items in the dataset. The mean grade of 5.41 and median of 5.70 place the central tendency just above 
the Fail--Pass boundary, with the median slightly higher than the mean indicating a mild negative skew. Grades span the full practical range of the scale (0.0--10), confirming that the dataset captures the complete spectrum of student performance. 
At question level, evaluators assigned, on average, the maximum possible mark on 557 individual responses and a zero on 350, reflecting that both fully correct and fully incorrect answers were frequent across the examined questions. Regarding the 
categorical distribution, 42.5\% of students received a Fail grade, while the remaining 57\% passed, distributed across Pass (26.4\%), Merit (25.3\%), and Distinction (5.3\%).

\subsubsection{Score Distribution}
\label{subsubsec:score_dist_humans}

\begin{figure}[h]
\centering
    \includegraphics[width=0.75\textwidth]{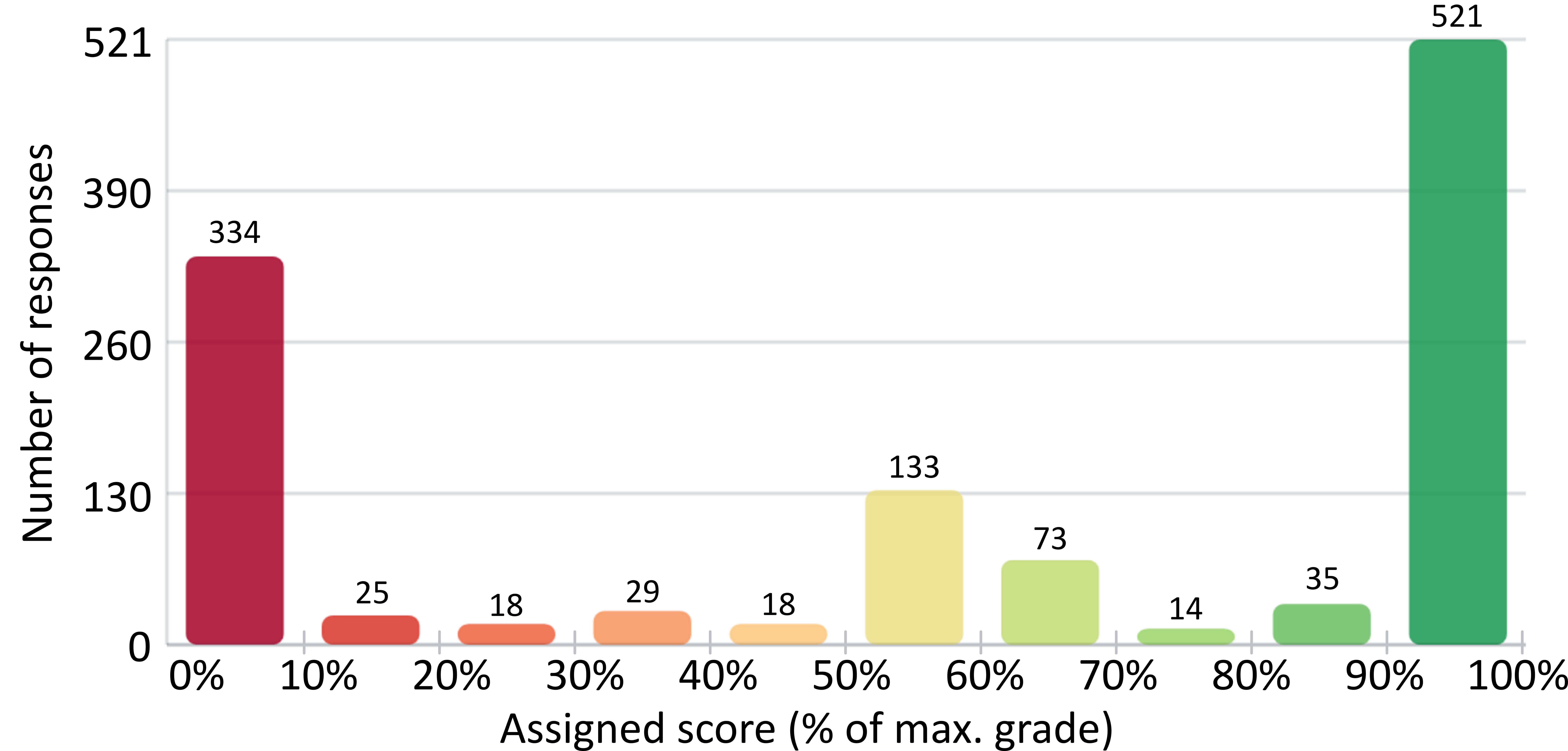}
    \caption{Frequency distribution of normalized item-level scores assigned by human evaluators.}
    \label{fig:distribucion_norm_humanos}
\end{figure}

Figure~\ref{fig:distribucion_norm_humanos} presents the frequency distribution of individual question level scores assigned by the human evaluators, expressed as the  proportion of the maximum achievable mark per question (0\% to 100\%, in decile bins).

The human distribution exhibits a strongly bimodal structure, with dominant peaks at the 0\% bin ($n = 334$) and the 90\% bin ($n = 521$), and a markedly lower  frequency in the intermediate deciles (10\%--80\%). This U-shaped profile indicates 
that human evaluators most frequently awarded either no credit or near-full credit  for individual items, with comparatively few outputs falling in the middle range. 
The 50\% bin constitutes the only notable intermediate concentration ($n = 133$), suggesting that partial credit was primarily awarded at the midpoint of the scale  rather than being distributed uniformly across the intermediate range. The 60\% bin 
presents a secondary local accumulation ($n = 73$), while the remaining intermediate deciles (10\%--40\% and 70\%--80\%) record negligible frequencies. 

This pattern is consistent with the nature of the assessment instrument: command-line responses are either functionally correct thus earning full or near-full marks, or substantially incorrect therefore earning zero or minimal credit with relatively few opportunities for finely graded partial solutions. The concentration at 50\%, however, reflects the rubric's provision for partial credit when a response demonstrates conceptual understanding despite syntactic or functional errors, a 
grading decision that requires the nuanced judgment characteristic of expert human evaluators.

\subsubsection{Assigned Score by Taxonomy Level}
\label{subsubsec:taxonomy_humans}

\begin{figure}[h]
\centering
    \includegraphics[width=0.7\textwidth]{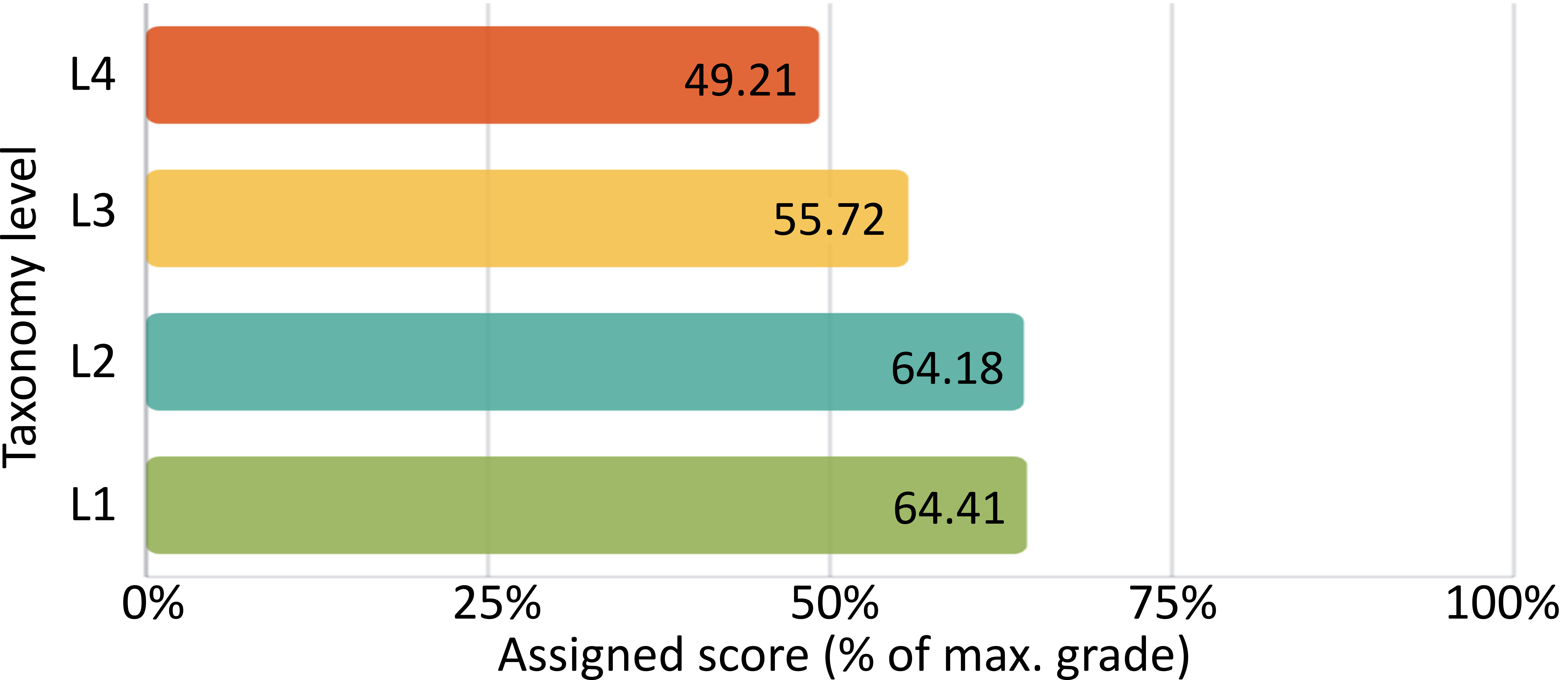}
    \caption{Distribution of assigned grade per taxonomy level for human evaluators.}
    \label{fig:stats_taxonomia_humans}
\end{figure}

Figure~\ref{fig:stats_taxonomia_humans} presents the mean normalized score per taxonomic level, expressed as the proportion of the maximum achievable mark. L1 yields the highest score (64.41\%), with L2 following extremely closely at 64.18\%. This near-tie is consistent with the expected difficulty ordering where simpler tasks yield a better performance.

L1 commands primarily demand exact recall: students must reproduce precise flag combinations, option names, and argument syntax, and the rubric deducts marks for missing flags, incorrect path specifications, or wrong argument ordering, meaning a conceptually correct answer can still incur penalties from minor syntactic gaps. L2 questions, by contrast, target the execution of learned procedures in concrete contexts, precisely the type of task that students rehearse most intensively through the practical exercises that accompany this course. Repeated action–consequence practice tends to consolidate this procedural knowledge robustly, partially offsetting the higher cognitive demand relative to L1. Performance declines clearly at L3 (55.72\%), reflecting the added difficulty of questions requiring structural understanding of permission models, data streams, and pattern-matching formalisms, and drops further at L4 (49.21\%), where approximately half of the available marks were achieved on average. This monotonic decrease from L2 onward is consistent with the increasing cognitive and operational demands described in the used taxonomy (Section~\ref{subsec:taxonomy}).

\subsubsection{Overall statistical results}
\label{subsubsec:stats_humans}

Table~\ref{tab:interrater} presents the comprehensive inter-rater reliability analysis among the three human evaluators. The results demonstrate exceptionally high agreement across all computed metrics, establishing a robust baseline for subsequent comparison with automated AI-based assessment.

\begin{table}[h]
\centering
\caption{Inter-rater reliability metrics.}
\begin{tabular}{lccc}
\hline
\textbf{Metric} & \textbf{Value} & \textbf{95\% CI}  \\
\hline
Pearson $r$ & 0.949 & [0.942, 0.955]  \\
Spearman $\rho$ & 0.943 & [0.934, 0.950]  \\
ICC(2,1) & 0.949 & [0.941, 0.955]  \\
Weighted $\kappa$ & 0.948 & [0.948, 0.954]  \\
MAE & 0.028 & --- \\
Bland-Altman Bias & -0.008 & [-0.010, -0.005] \\
\hline
\end{tabular}\label{tab:interrater}
\end{table}

The correlation analyses reveal near-perfect linear association between evaluators' scores. Pearson's correlation coefficient of $r = 0.949$ (95\% CI $[0.942, 0.955]$, $p < 0.001$) indicates that evaluators maintained highly consistent proportional scoring patterns across the full range of student performance levels. Similarly, Spearman's rank correlation of $\rho = 0.943$ (95\% CI $[0.934, 0.950]$, $p < 0.001$) confirms that evaluators achieved nearly identical rank-ordering of students, demonstrating consistency not only in absolute scores but also in relative performance assessment. The minimal difference between Pearson and Spearman coefficients suggests that the linear relationship is maintained throughout the scoring distribution without substantial influence from outliers or non-linear patterns.

The Intraclass Correlation Coefficient $\text{ICC}(2,1) = 0.949$ (95\% CI $[0.941, 0.955]$, $p < 0.001$) indicates excellent absolute agreement. This metric is particularly informative as it quantifies the proportion of total variance attributable to true differences in student performance rather than evaluator inconsistency. An ICC(2,1) above 0.90 signifies that more than 90\% of the observed score variance reflects genuine differences in student proficiency, with less than 10\% attributable to measurement error or evaluator disagreement. This level of reliability exceeds the threshold typically required for high-stakes educational assessment
contexts.

Agreement metrics further corroborate the exceptional consistency between evaluators. Weighted Kappa reached $\kappa_w = 0.948$ (95\% CI $[0.948, 0.954]$), a value indicating that the disagreements that did occur were predominantly minor in magnitude since quadratic weighting would substantially penalise larger mismatches if they were frequent.

The Mean Absolute Error (MAE) of $0.028$ points provides an intuitive measure of typical scoring discrepancy expressed in the original grade scale. This low value indicates that, on average, evaluators' scores differed by less than three hundredths of a point, which is a practically negligible difference in educational assessment contexts. Given that individual question scores
ranged from 0.25 to 1.25 points, this level of precision demonstrates exceptional calibration between evaluators.

Bland-Altman analysis revealed no significant systematic bias between evaluators. The mean difference of $-0.008$ points (95\% CI $[-0.010, -0.005]$) indicates that neither evaluator consistently assigned higher or lower scores than the other. This near-zero bias confirms that the observed trivial difference could readily be attributed to random variation rather than
systematic scoring tendencies, providing strong evidence that the three evaluators share a common scoring philosophy.

These results establish that the three human evaluators achieved exceptional consistency in their independent assessments of student responses. The combination of high correlation, excellent ICC values, near-perfect agreement coefficients, minimal absolute error, and absence of systematic bias provides strong evidence that the human evaluation represents a reliable
and valid reference standard against which to compare automated AI-based assessment methods.

\subsection{LLM Evaluators Results}\label{subsec:llm_results}

This section presents the grading outputs produced by the four LLMs under both prompt variants, namely Variant 1 - No Rubric (V1) and Variant 2 - With Rubric (V2), as described in Section~\ref{subsubsec:eval_process}. The distribution of item-level scores  is first examined to characterize how each model allocates marks across the response quality spectrum (Section~\ref{subsubsec:assigned_score}). Then, performance is stratified by taxonomy level to identify where agreement with human evaluation is strongest and where it deteriorates (Section~\ref{subsubsec:taxonomy_level}). Finally, aggregate descriptive statistics and grade-band distributions are reported to enable direct comparison with the human baseline established in Section~\ref{subsec:human_consensus}. 

\subsubsection{Score Distribution}\label{subsubsec:assigned_score}

\begin{figure}[h]
\centering
\begin{subfigure}[b]{0.45\textwidth}
    \includegraphics[width=\textwidth]{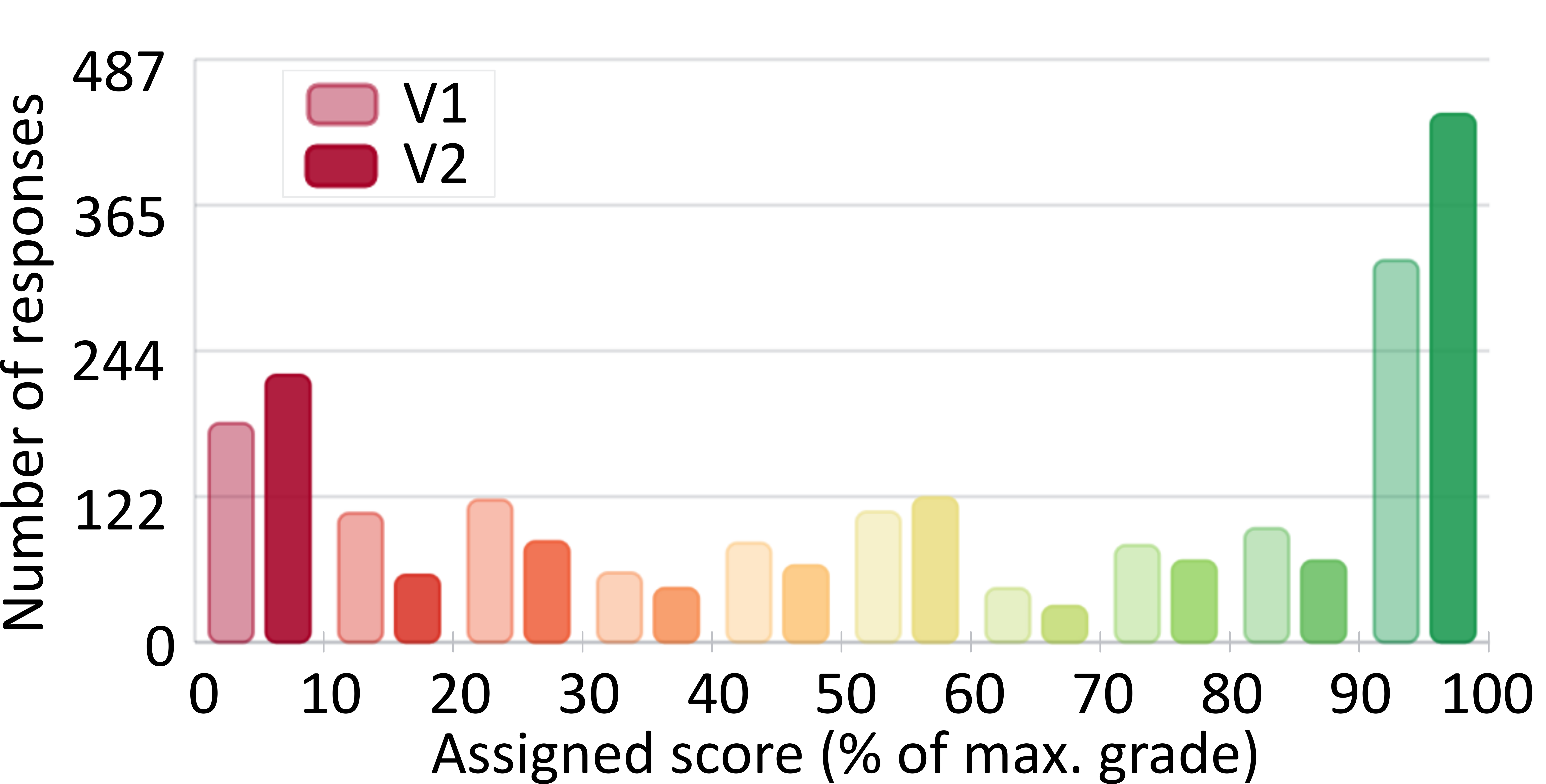}
    \caption{GPT-5.2.}
    \label{fig:variant1_gpt}
\end{subfigure}
\hfill
\begin{subfigure}[b]{0.45\textwidth}
    \includegraphics[width=\textwidth]{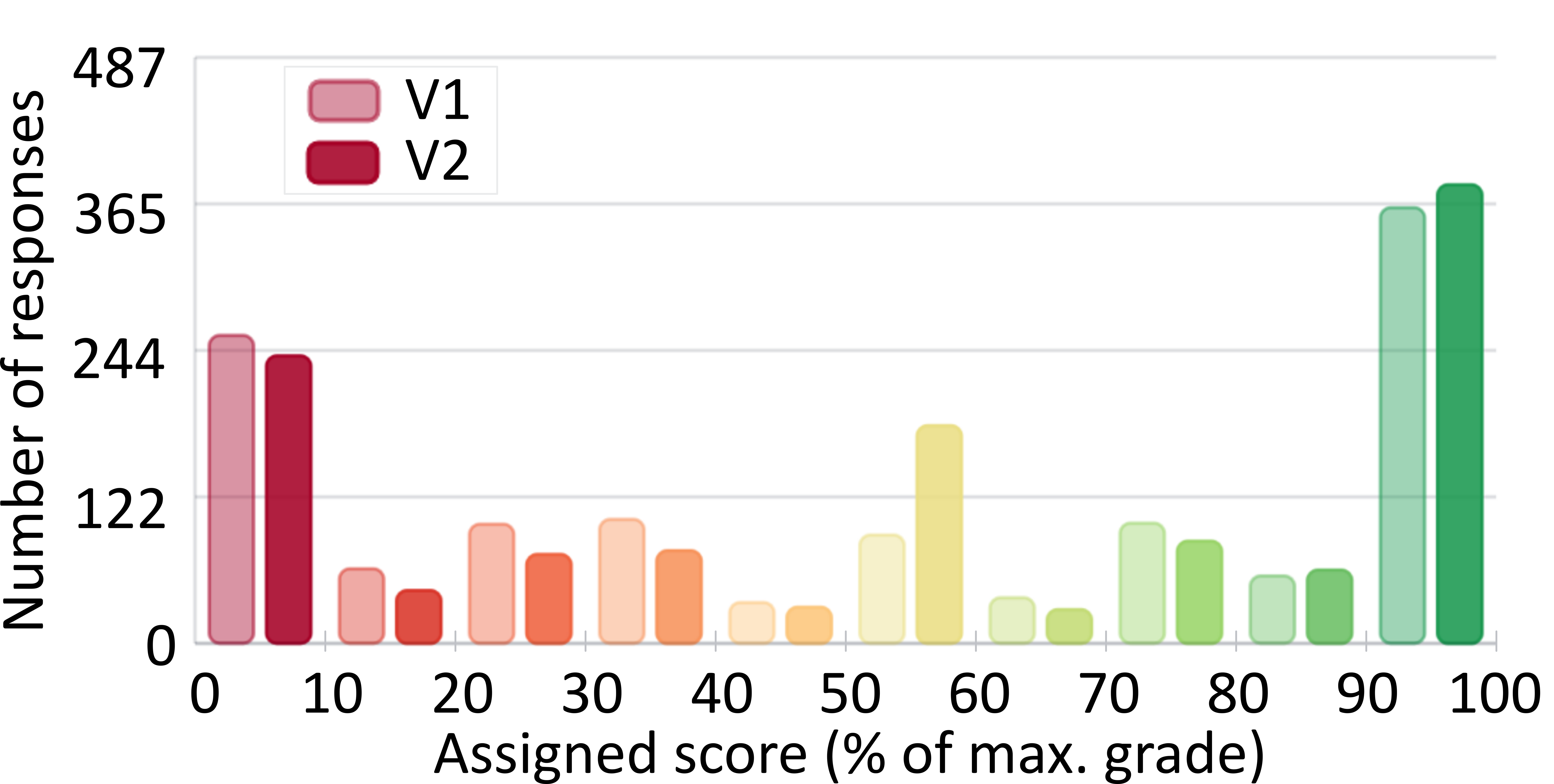}
    \caption{Claude Opus 4.6.}
    \label{fig:variant1_claude}
\end{subfigure}

\begin{subfigure}[b]{0.45\textwidth}
    \includegraphics[width=\textwidth]{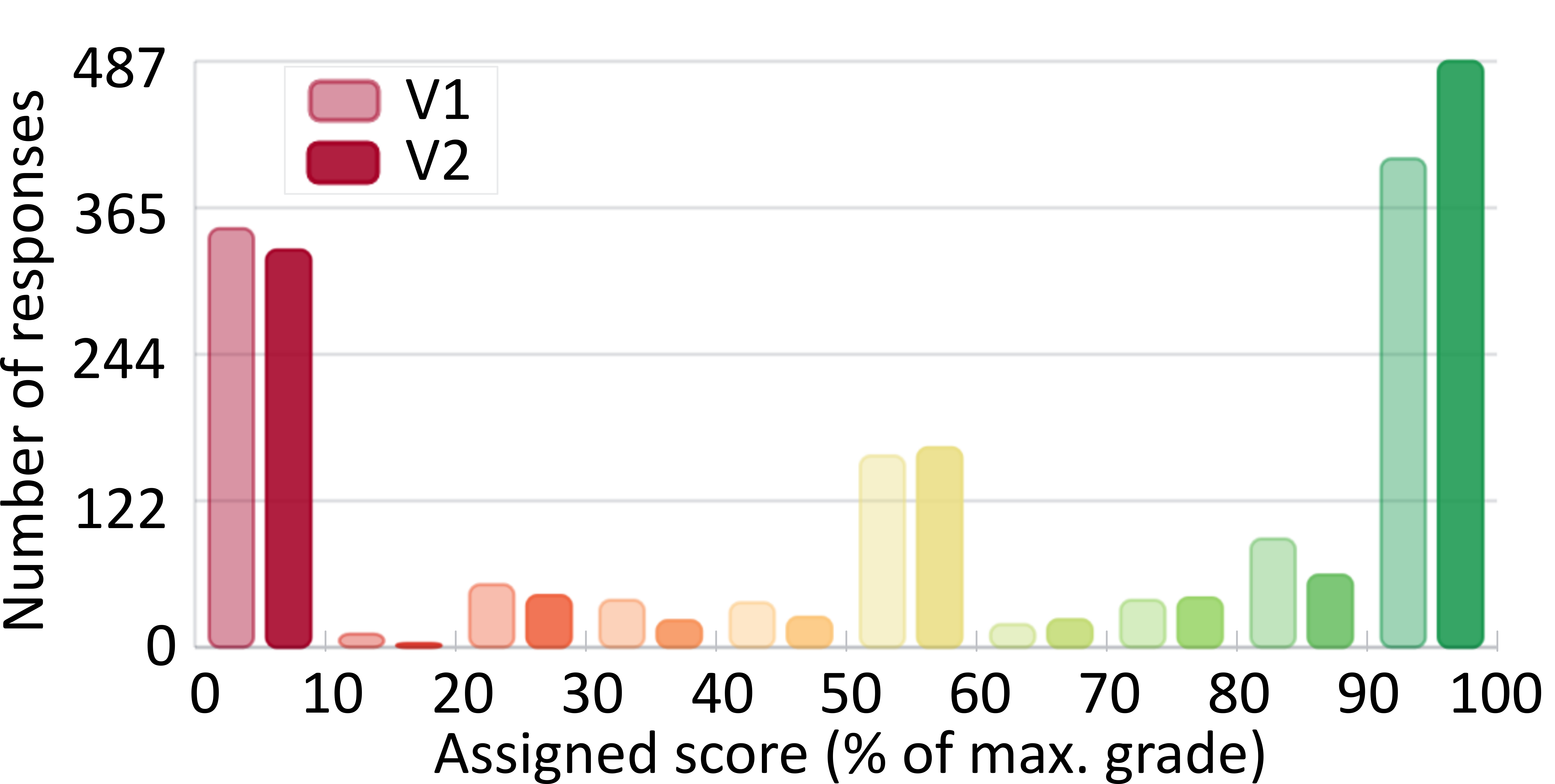}
    \caption{Gemini 3.0 Pro.}
    \label{fig:variant1_gemini}
\end{subfigure}
\hfill
\begin{subfigure}[b]{0.45\textwidth}
    \includegraphics[width=\textwidth]{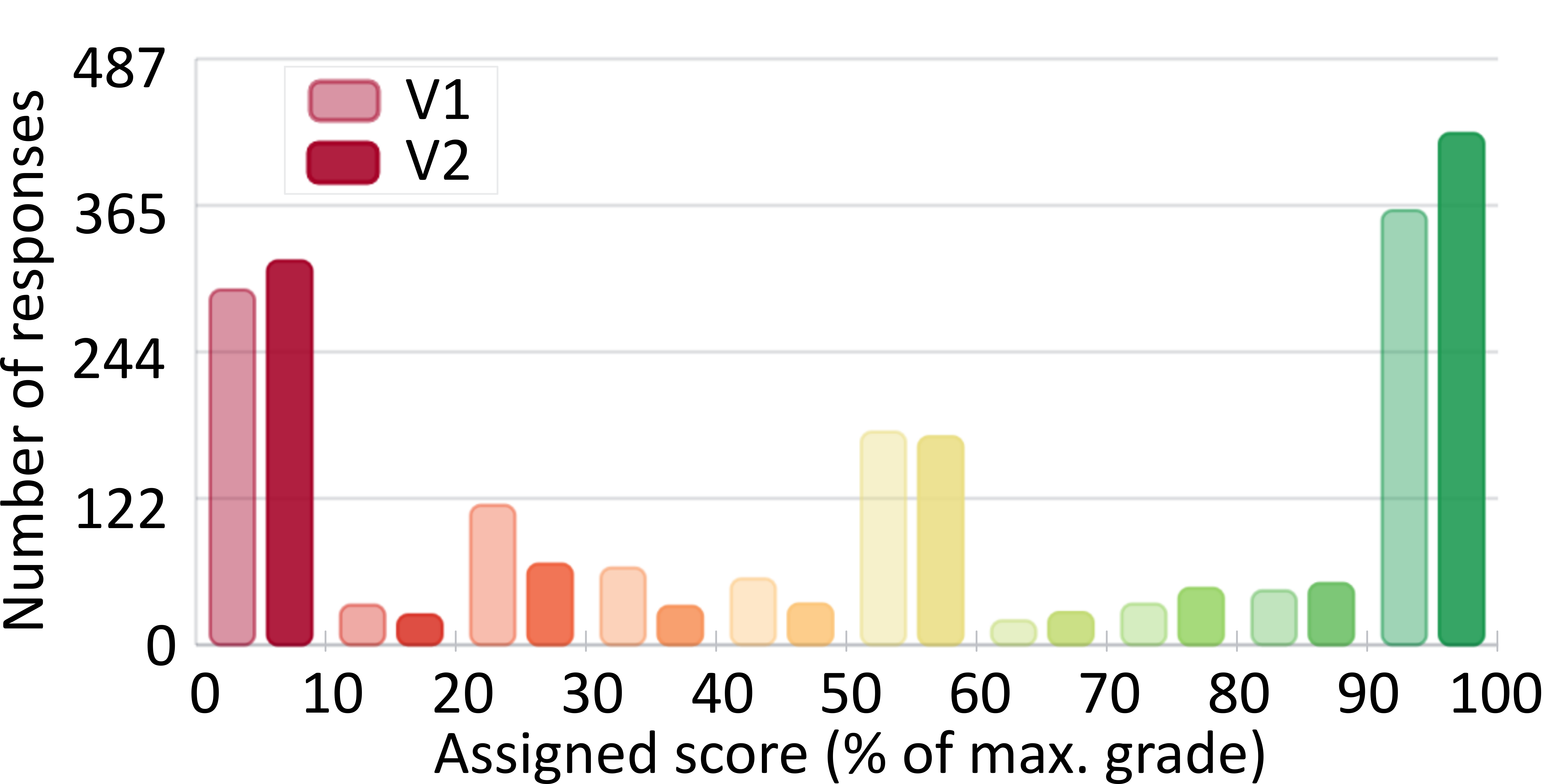}
    \caption{GLM 5.}
    \label{fig:variant1_glm}
\end{subfigure}
\caption{Distribution of assigned grade per question for Variant 1 and Variant 2 across LLM models.}
\label{fig:v1_vs_v2_grade_distributions}
\end{figure}

Figure~\ref{fig:v1_vs_v2_grade_distributions} presents the frequency distributions of individual item-level scores. Within each chart, both variants are shown, with the lighter colour representing V1 and the darker colour representing V2. 

GPT displays a broad, relatively flat distribution across the full range, with no single dominant intermediate peak. The 0\% bin is substantial in both variants (V1=183, V2=223), and the 90\% bin is the tallest across the entire chart (V1=319, V2=441), indicating a strong tendency toward near-perfect scores. The intermediate bins (10\%–80\%) are broadly distributed, with generally comparable frequencies between variants. The V2 variant increases the 90\% bin substantially while also increasing the 0\% bin slightly, suggesting that GPT V2 polarizes the distribution further toward the extremes. A local peak at 50\% is present in V2 (121) but not in V1 (109), indicating a moderate additional concentration of mid-range outputs.

Claude shows a profile similar in structure to GPT, with dominant peaks at 0\% (V1=256, V2=239) and 90\% (V1=362, V2=381), and a secondary local concentration around 50\% that is more pronounced in V2 (181) than in V1 (90). The transition from V1 to V2 in Claude is characterized primarily by a redistribution of mass toward the 50\% bin, accompanied by a modest increase in the 90\% bin and a slight reduction in the 0\% bin. This pattern suggests that Claude V2 improves a subset of zero-scored outputs to the mid-range level, while also incrementally increasing the proportion of near-perfect outputs.

Gemini exhibits the most distinctive distributional profile among all four models: a pronounced U-shaped pattern, with very high frequencies at both the 0\% bin (V1=348, V2=330) and the 90\% bin (V1=406, V2=487), and substantially lower, relatively flat frequencies in the intermediate deciles (10\%-80\%). This bimodal structure indicates that Gemini produces a large proportion of near-zero scores alongside a large proportion of near-perfect scores, with comparatively few outputs in the middle range. The V2 variant amplifies this pattern: the 90\% bin grows substantially while the 0\% bin decreases modestly, suggesting that V2 converts some low-scoring outputs into high-scoring ones rather than improving scores uniformly across the distribution.

GLM presents a strongly bimodal distribution, with dominant peaks at 0\% (V1=295, V2=319) and 90\% (V1=361, V2=425), and a secondary local concentration around 50\% (V1=177, V2=173). Unlike what was previously reported, the transition from V1 to V2 in GLM does not reduce the zero-score bin. Both extreme bins increase under V2, with the most substantial growth occurring at 90\% (+85). The intermediate deciles remain relatively flat and low in both variants, with the partial exception of 20\% (116), which decreases slightly under V2 (67). This pattern aligns GLM more closely with Gemini and GPT in terms of its overall U-shaped profile. The V2 in GLM marks an amplification of the high-scoring.

All four models share a common structural feature: a strongly right-skewed distribution with a dominant peak at 90\% and a secondary at 0\%, indicating that item-level scoring is inherently bimodal in nature. Outputs are either answered very well or not at all. The intermediate deciles (10\%-80\%), with the exception of 50\% bin, consistently show low absolute frequencies relative to the extremes, regardless of model or variant. The V2 variants generally amplify the 90\% peak while modestly reducing the 0\% bin. Gemini exhibits the most extreme bimodality, while GPT shows the largest absolute increase in the 90\% bin under V2. Claude is distinguished by the largest relative increase in the 50\% scoring between variants, suggesting a distinct mechanism of improvement compared to the other three models.

\subsubsection{Assigned Score by Taxonomy Level}\label{subsubsec:taxonomy_level}

\begin{figure}[h]
\centering
\begin{subfigure}[b]{0.45\textwidth}
    \includegraphics[width=\textwidth]{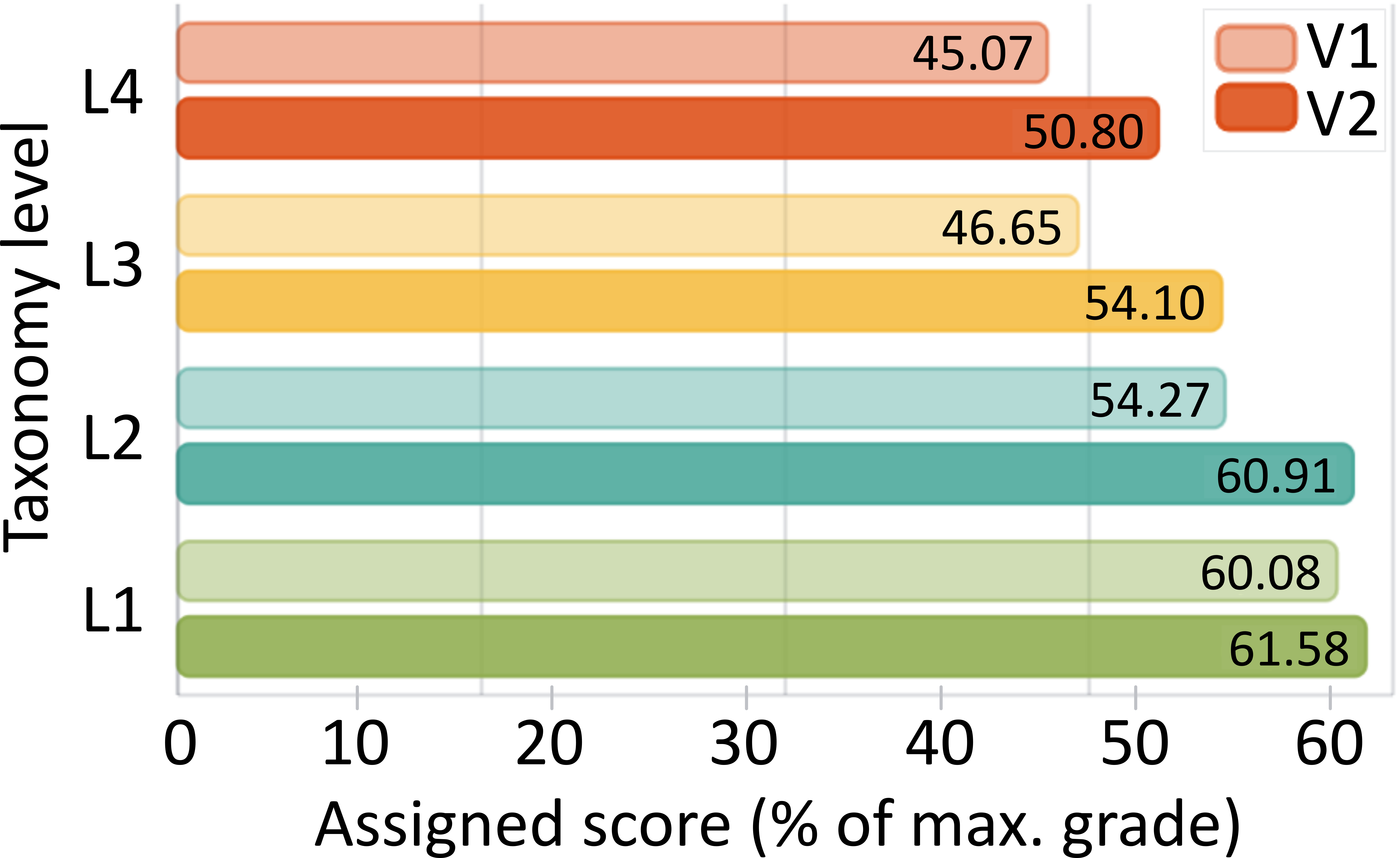}
    \caption{GPT-5.2.}
    \label{fig:nota_tax_gpt}
\end{subfigure}
\hfill
\begin{subfigure}[b]{0.45\textwidth}
    \includegraphics[width=\textwidth]{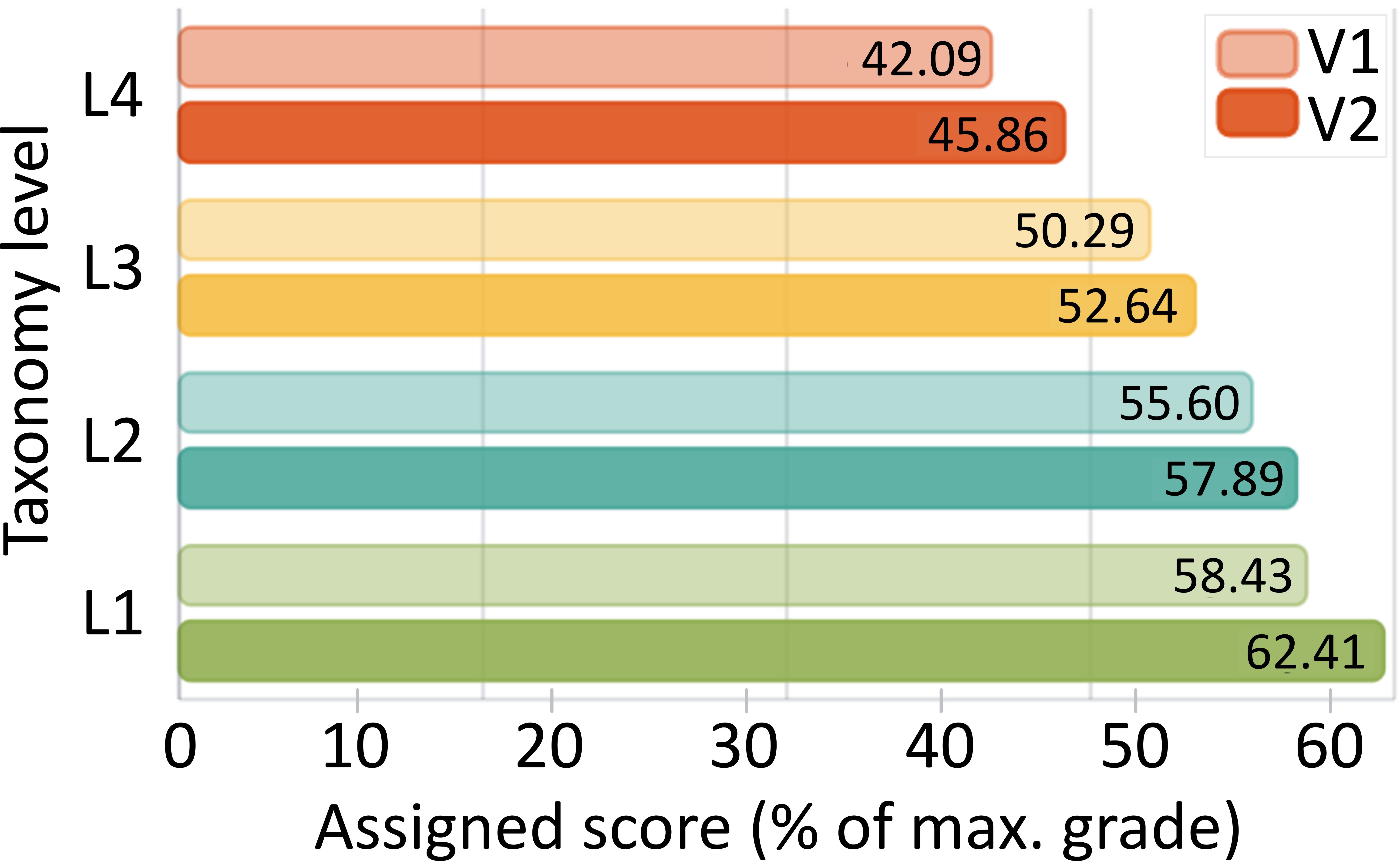}
    \caption{Claude Opus 4.6.}
    \label{fig:nota_tax_claude}
\end{subfigure}

\begin{subfigure}[b]{0.45\textwidth}
    \includegraphics[width=\textwidth]{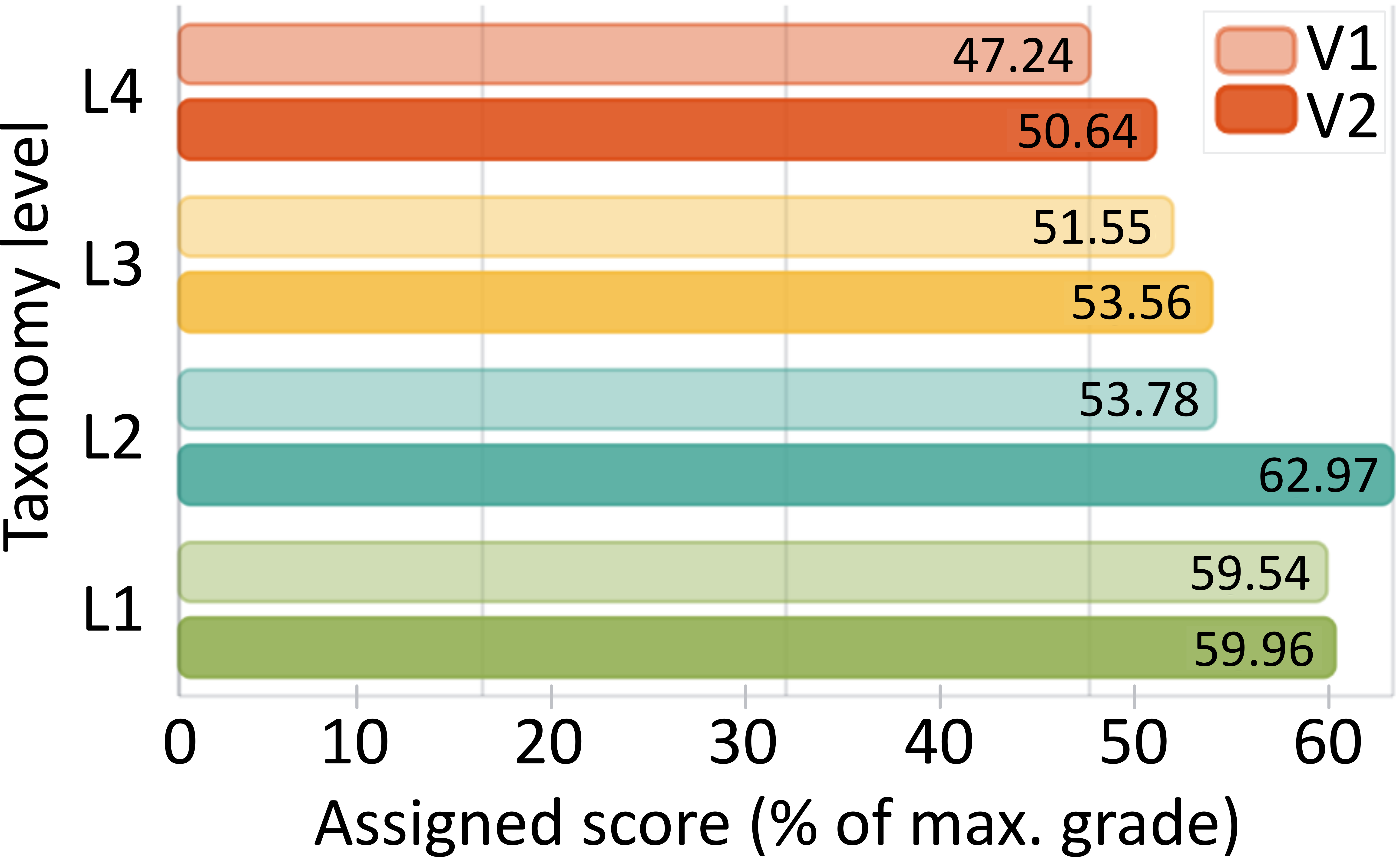}
    \caption{Gemini 3.0 Pro.}
    \label{fig:nota_tax_gemini}
\end{subfigure}
\hfill
\begin{subfigure}[b]{0.45\textwidth}
    \includegraphics[width=\textwidth]{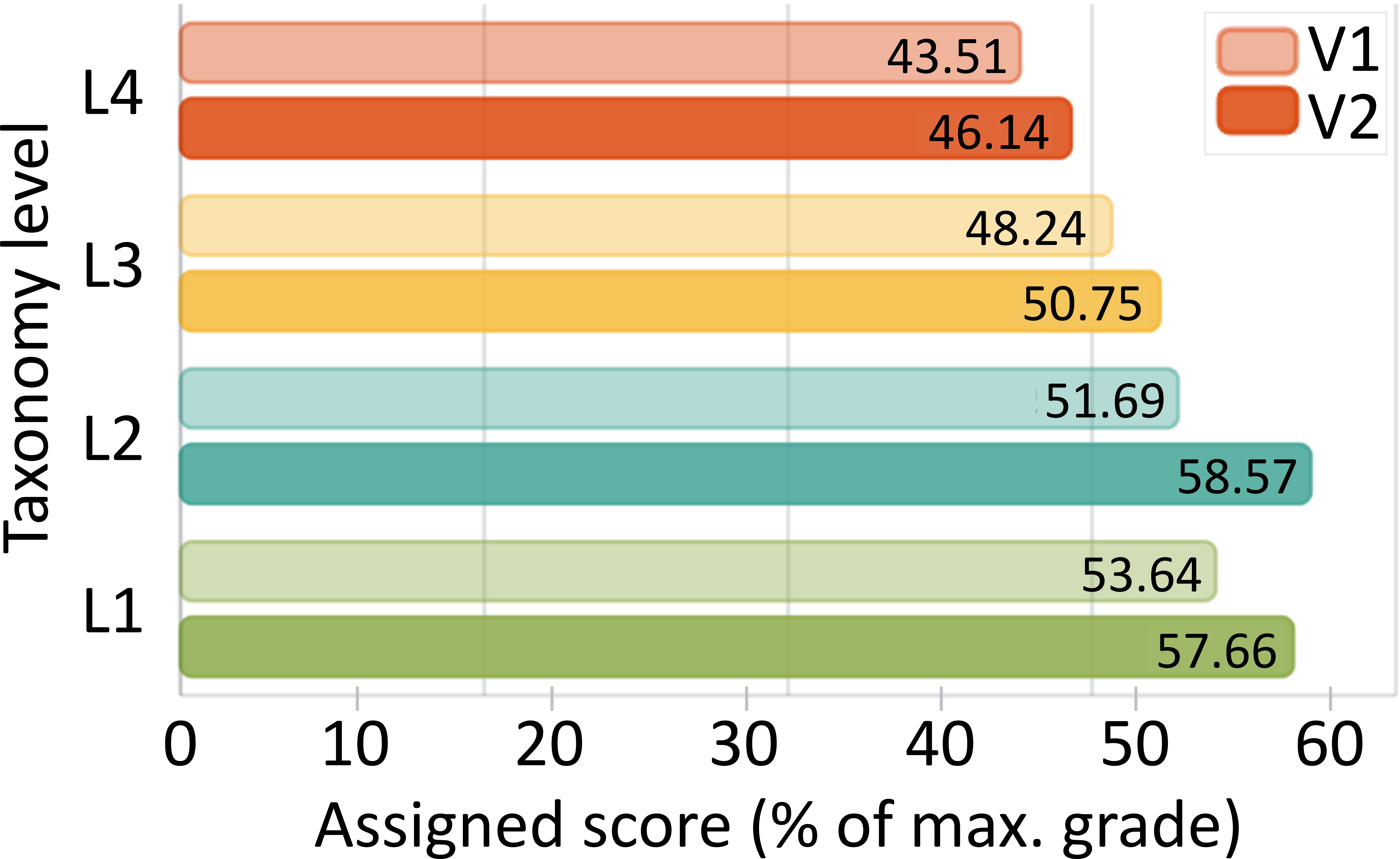}
    \caption{GLM 5.}
    \label{fig:nota_tax_glm}
\end{subfigure}
\caption{Distribution of assigned grade per taxonomy level for Variant 1 and Variant 2 across LLM models.}
\label{fig:v1_vs_v2_taxonomy_distributions}
\end{figure}

Figure~\ref{fig:v1_vs_v2_taxonomy_distributions} presents the performance scores obtained by GPT, Claude, Gemini, GLM evaluated across the four-level cognitive taxonomy. Scores range from 0\% to 100\%, with higher values indicating the percentage of mark associated to the different questions of each category. 

GPT maintains a consistent positive gap in favour of V2 across all taxonomy levels, with the largest advantage observed at L3 (approximately 7.5 percentage points). At L4, questions received on average 50.80\% of the available mark under V2 and 45.07\% under V1, the highest proportions recorded at the this level among all four models.

Claude assigns an average of 62\% of the available mark at L1 under V2 and 58.43\% under V1, with scores declining progressively to 45.86\% (V2) and 42.09\% (V1) at L4. The inter-variant gap remains relatively stable across all taxonomy levels (approximately 3 percentage points), indicating consistent but moderate gains from V2 regardless of cognitive complexity.

Gemini shows a notable difference at L2, where questions receive on average 62.97\% of the available mark under V2 (the highest proportion at that level among all four models) compared to 53.78\% under V1. This unusually large inter-variant gap (approximately 9 percentage points) is not observed at adjacent levels, suggesting that V2 introduces a specific improvement in the scoring of basic-level tasks. At L4, the two variants converge, with questions receiving 59\% of the available mark.

GLM assigns similar score proportions to previous models. The monotonic decline toward higher cognitive levels follows the general pattern observed across all models: at L3, questions receive on average 50.75\% of the available mark under V2, falling further to 46.14\% at L4. The inter-variant gap at L4 stands at 2.63 percentage points.

A consistent pattern emerges across all four models: the proportion of the available mark assigned to questions decreases monotonically as cognitive complexity increases. Questions at the L1 level receive the highest score proportions across all models and variants, while those at the L4 level receive the lowest, indicating that all evaluated models award a smaller share of the available mark as tasks shift from retrieval-based recall toward advanced reasoning and synthesis.
At L4, GPT V2 and Gemini V2 assign the highest score proportions (50.80\% and 50.64\% respectively), while Claude V2 (45.86\%) and GLM V2 (46.14\%) trail behind, suggesting that GPT and Gemini maintain a relative advantage in allocating marks to high-complexity tasks. At L1, inter-model differences are smaller, with all V2 configurations falling within a 3-percentage-point range, with the exception of Gemini and GLM, whose highest score proportion occurs at L2 rather than L1.

Across all models and taxonomy levels, V2 consistently assigns a higher proportion of the available mark than V1, though the magnitude of the difference varies. The largest inter-variant gaps are observed at L2 for Gemini and at L3 for GPT, suggesting that the changes introduced in V2 yield the most pronounced improvements at intermediate levels of cognitive complexity rather than at the extremes of the taxonomy.

\subsubsection{Overall statistical results}

The descriptive statistics presented in Table~\ref{tab:variant1_descriptive} summarise the final grade distributions produced by the four large language models. For each model–variant combination, the table reports mean and median final grade, observed score range, item-level counts of perfect (100\%) and zero (0\%) scores, and the proportion of evaluation instances falling within four grade bands.

\begin{table}[h]
\centering
\caption{Descriptive statistics of LLM final grades. The first column shows the LLM model and the prompt variant (V1 and V2)}
\label{tab:variant1_descriptive}
\small
\begin{tabular}{lccccccccc}
\toprule
\textbf{Model [V]} & \textbf{Mean} & \textbf{Median} & \textbf{Min/Max} & \textbf{Perfect} & \textbf{Zeros} & \textbf{Fail} & \textbf{Pass} & \textbf{Merit} & \textbf{Dist.} \\
\midrule
Claude  [V1] & 4.73 & 4.95 & 0.6\,/\,9.5 & 341 & 245 & 50.6\% & 40.2\% & 8.0\% & 1.1\% \\
Claude  [V2] & 5.05 & 5.25 & 0.7\,/\,9.4 & 367 & 232 & 47.1\% & 29.9\% & 21.8\% & 1.1\% \\
\midrule
Gemini  [V1] & 4.98 & 5.50 & 0.3\,/\,9.3 & 371 & 346 & 44.8\% & 34.5\% & 18.4\% & 2.3\% \\
Gemini  [V2] & 5.39 & 5.85 & 0.4\,/\,9.8 & 452 & 327 & 41.4\% & 27.6\% & 28.4\% & 2.3\% \\
\midrule
GPT     [V1] & 4.84 & 5.21 & 0.5\,/\,8.2 & 260 & 172 & 45.9\% & 45.9\% & 8.0\% & 0.0\% \\
GPT     [V2] & 5.40 & 5.70 & 0.3\,/\,9.4 & 381 & 213 & 37.9\% & 32.2\% & 28.7\% & 1.1\% \\
\midrule
GLM     [V1] & 4.61 & 4.89 & 0.3\,/\,9.1 & 341 & 292 & 52.9\% & 34.5\% & 11.5\% & 1.1\% \\
GLM     [V2] & 5.00 & 5.10 & 0.4\,/\,9.1 & 402 & 312 & 48.3\% & 27.6\% & 21.8\% & 2.3\% \\
\bottomrule
\end{tabular}
\end{table}

All mean final grades fall within a narrow range of approximately 0.8 points, from 4.61 (GLM, V1) to 5.40 (GPT, V2). In every configuration, the median exceeds the corresponding mean, indicating a consistent negative skew across models and variants. The V2 variant yields higher mean and median grades than V1 for all four models, with mean improvements ranging from 0.32 points (Claude) to 0.56 points (GPT). In all configurations, the gap between the lowest and highest final grade assigned spans approximately 8--9 points, indicating that switching from V1 to V2 shifts grades upward on average but does not narrow the spread between the best and worst scoring students. The sole exception to this variability pattern is GPT V1, which presents the most constrained maximum observed grade (8.2) and is the only configuration in which no evaluation instance reached the Distinction threshold; this ceiling is absent in GPT V2, where the maximum rises to 9.4.

The counts of item-level perfect and zero scores provide complementary evidence regarding the mechanism underlying grade improvements between variants. Under V2, all four models increase their count of perfect-scoring items, with gains ranging from 26 (Claude) to 121 (GPT). The behaviour of zero scores is, however, asymmetric across models: Claude and GPT reduce their zero counts by 13 and 59 items respectively, whereas Gemini and GLM record modest increases. This divergence suggests that improvement is not uniformly achieved through the conversion of zero-scoring items into positive-scoring ones. For Claude and GPT, V2 appears to reduce outright scoring failures at the question level, while for Gemini and GLM the improvement in final grades is driven primarily by an increase in perfect-scoring items rather than by a reduction in zeros.

Turning to the grade band distributions, the Fail rate constitutes the dominant category across all V1 configurations, ranging from 44.8\% (Gemini) to 52.9\% (GLM), indicating that the majority of evaluation instances did not reach the passing threshold under the first variant. V2 reduces the Fail rate for all models, with the largest decrease observed for GPT and the smallest for Claude. The Pass band contracts under V2 in all models despite the general upward shift in grades, a result explained by the simultaneous and substantial expansion of the Merit band. Merit rates increase by 20.7\% for GPT, 13.8\% for Claude, 10.3\% for GLM, and 10.0\% for Gemini. This pattern indicates that the primary focus of improvement between variants lies at the Pass--Merit boundary: V2 disproportionately elevates instances that were already near the passing threshold into the Merit category, rather than producing a uniform upward shift across the full grade distribution. The Distinction rate remains negligible in all configurations, reaching a maximum of 2.3\% in three cases (Gemini V1, Gemini V2, GLM V2) and 0.0\% for GPT V1, suggesting that near-perfect final grades are rarely achieved under the current evaluation framework regardless of model or variant.

In terms of cross-model comparison, Gemini and GPT achieve the highest V2 means (5.39 and 5.40 respectively) and the highest Merit rates under V2 (28.4\% and 28.7\%). Claude and GLM trail marginally, with V2 means of 5.05 and 5.00. Nevertheless, GPT exhibits the most structurally changed profile between variants. It records the largest absolute increase in perfect scores (+121), the largest reduction in zero scores (-59), the largest decrease in Fail rate (-8.0\%), and the largest increase in Merit rate (+20.7\%). Taken together, these results indicate that while all four models benefit from the V2 variant, the nature and magnitude of improvement vary considerably across models, and that aggregate measures such as mean grade alone are insufficient to characterize the distributional shifts induced by the variant change.

Collectively, the results presented in this section establish three empirical regularities that will guide the subsequent discussion. First, all four LLMs produce grade distributions and central tendencies broadly consistent with the human baseline under rubric-guided prompting, yet none fully replicates the distributional shape of expert evaluation, particularly at the upper end of the grade spectrum. Second, the introduction of a structured rubric (V2) yields consistent and non-trivial improvements in human--AI agreement across all models and metrics. Third, agreement degrades systematically as cognitive complexity increases, with the largest discrepancies observed at taxonomy levels L3 and L4. The following section interprets these patterns in terms of their practical and theoretical implications for AI-assisted assessment.

\section{Discussion}
\label{sec:discussion}

The empirical evidence reported in the previous section enables a structured examination of the conditions under which LLM-assisted grading approximates expert human evaluation. Rather than asking whether automated assessment is feasible in the abstract, the data allow more precise questions: which models, under which prompting conditions, and at which levels of cognitive complexity can automated scoring be trusted? To address these questions, this section first quantifies human--AI agreement through a comprehensive battery of statistical metrics, then stratifies that agreement by taxonomy level to expose where the limits of automated evaluation lie, and finally draws out the theoretical implications of the observed patterns.

Table~\ref{tab:human_ai_agreement} presents comprehensive agreement metrics quantifying the concordance between LLM-generated grades and the human evaluator consensus across all 1200 student responses. The results demonstrate substantial but imperfect alignment, with all models falling measurably short of the human inter-rater reliability ceiling established in Section~\ref{subsec:human_consensus} (ICC(2,1) = 0.949, $\kappa_w$ = 0.948, Pearson $r$ = 0.949).

\begin{table}[h]
\centering
\caption{Human--AI agreement metrics across LLM models and variants. Bold values indicate the best result per metric.}
\label{tab:human_ai_agreement}
\small
\begin{tabular}{lcccccc}
\toprule
\textbf{Model [V]} & \textbf{Pearson} & \textbf{Spearman} & \textbf{ICC} & \textbf{Weighted $\kappa$} & \textbf{MAE} & \textbf{Bland-Altman} \\
\midrule
Claude [V1]     & 0.811 & 0.808 & 0.809 & 0.796 & 0.165 & $-0.07$  \\
Claude [V2]   & 0.867 & 0.862 & 0.863 & 0.859 & 0.124 & $-0.039$ \\
\midrule
Gemini [V1]     & 0.829 & 0.811 & 0.829 & 0.821 & 0.144 & $-0.057$ \\
Gemini [V2]   & \textbf{0.888} & \textbf{0.877} & \textbf{0.888} & \textbf{0.886} & \textbf{0.100} & \textbf{-0.014} \\
\midrule
GLM [V1]        & 0.795 & 0.798 & 0.793 & 0.773 & 0.170 & $-0.093$ \\
GLM [V2]      & 0.845 & 0.838 & 0.844 & 0.838 & 0.126 & $-0.051$ \\
\midrule
GPT [V1]        & 0.787 & 0.780 & 0.779 & 0.767 & 0.184 & $-0.07$  \\
GPT [V2]      & 0.850 & 0.843 & 0.847 & 0.845 & 0.132 & $-0.016$ \\
\bottomrule
\end{tabular}
\end{table}

Across all metrics, Gemini V2 achieves the highest human--AI agreement: Pearson correlation of 0.888, ICC(3,1) of 0.888, and Weighted Kappa of 0.886. These values indicate strong linear association and excellent rank-order preservation, approaching, but not matching, the human baseline. The Mean Absolute Error of 0.100 points represents the lowest typical scoring discrepancy among all configurations, corresponding to one-tenth of a grade point per question. The Bland-Altman bias of $-0.014$ reveals minimal systematic tendency, indicating that Gemini V2 neither consistently over-grades nor under-grades relative to human evaluators.

The rubric-enhanced variant (V2) consistently outperforms the baseline prompt (V1) across all four models and all metrics. The improvement is particularly pronounced for Gemini (ICC(3,1) increases from 0.829 to 0.888, $+$0.059) and Claude (ICC(3,1) increases from 0.809 to 0.863, $+$0.054), demonstrating that structured evaluation criteria substantially enhance concordance with expert judgment. GPT exhibits the largest MAE reduction from V1 to V2 (from 0.184 to 0.132, $-$0.052), indicating that rubric guidance most effectively reduces scoring errors for this model. Bland-Altman bias also shifts closer to zero under V2 for all models, with Gemini V2 ($-0.014$) and GPT V2 ($-0.016$) achieving near-zero systematic bias.

Model-specific performance varies considerably. GLM records the lowest agreement in both variants (V1: ICC = 0.793; V2: ICC = 0.844), with the highest MAE (0.170 in V1) and most pronounced negative bias ($-0.093$ in V1), suggesting a systematic tendency toward under-grading relative to human evaluators. GPT V1 exhibits the poorest initial performance (ICC = 0.779, MAE = 0.184), but demonstrates the largest relative improvement under rubric guidance, reaching ICC = 0.847 in V2. Claude maintains intermediate performance in both variants, with V2 achieving ICC = 0.863 and MAE = 0.124, positioned between GLM and the top-performing Gemini configuration.

The convergence of Pearson, Spearman, and ICC values within each model--variant combination indicates that the relationships between human and LLMs scores are predominantly linear and monotonic, without substantial influence from outliers or rank reversals. This consistency across correlation types suggests that models maintain coherent scoring patterns across the full performance spectrum, rather than exhibiting erratic behavior at specific grade ranges.

Table~\ref{tab:icc_by_taxonomy} breaks down the level of agreement between human evaluators and the LLM across the different taxonomic levels, showing that this agreement varies systematically with cognitive complexity. This perspective helps identify the contexts in which automated assessment is more dependable and those in which human supervision is still necessary.

\begin{table}[h]
\centering
\caption{ICC(3,1) agreement stratified by taxonomy level across LLM models and variants.  Bold values indicate the best result per model.}
\label{tab:icc_by_taxonomy}
\small
\begin{tabular}{lccccc}
\toprule
\textbf{Model [V]} & \textbf{Global} & \textbf{L1} & \textbf{L2} & \textbf{L3} & \textbf{L4} \\
\midrule
Claude [V1]     & 0.809 & 0.838 & 0.773 & 0.791 & 0.820 \\
Claude [V2]   & 0.863 & \textbf{0.894} & 0.836 & 0.849 & 0.865 \\
\midrule
Gemini [V1]     & 0.829 & 0.871 & 0.762 & 0.844 & 0.853 \\
Gemini [V2]   & \textbf{0.888} & \textbf{0.894} & \textbf{0.888} & \textbf{0.880} & \textbf{0.883} \\
\midrule
GLM [V1]       & 0.793 & 0.827 & 0.712 & 0.795 & 0.852 \\
GLM [V2]      & 0.844 & 0.858 & 0.814 & 0.850 & 0.853 \\
\midrule
GPT [V1]        & 0.779 & 0.800 & 0.753 & 0.728 & 0.820 \\
GPT [V2]      & 0.847 & 0.833 & 0.872 & 0.846 & 0.820 \\
\bottomrule
\end{tabular}
\end{table}

At L1, agreement is highest and most uniform across models. Claude V2 and Gemini V2 both achieve ICC = 0.894, the maximum observed value at any taxonomy level for any model. GPT V2 (0.833) and GLM V2 (0.858) record lower but still strong agreement at L1. This convergence at the simplest cognitive level confirms that all models can reliably evaluate questions requiring exact recall of command syntax, flag combinations, and standard option patterns which are tasks where correctness is largely binary and rubric application is straightforward.

Performance diverges more substantially at L2. Gemini V2 achieves ICC(3,1) = 0.888, the highest L2 value and notably higher than its own L1 performance, suggesting that the rubric-enhanced prompt is particularly effective for procedural tasks involving file manipulation and directory navigation. GPT V2 also records strong L2 agreement (0.872), while Claude V2 (0.836) and GLM V2 (0.814) trail. In contrast, the baseline variants exhibit markedly lower L2 agreement, with GLM V1 recording ICC = 0.712, the lowest value at any level for any model except GPT V1 at L3.

At L3, agreement deteriorates for all models, though the magnitude of degradation varies. Gemini V2 maintains ICC = 0.880, demonstrating relative robustness at this complexity tier. GPT V2 (0.846), Claude V2 (0.849), and GLM V2 (0.850) cluster within a narrow band (0.846--0.850), indicating comparable difficulty in evaluating questions requiring understanding of permission models, data streams, and pattern-matching formalisms. The baseline variants exhibit more severe degradation, with GPT V1 recording ICC = 0.728 at L3, the lowest structural-level agreement observed.

At L4, the pattern shifts. Gemini V2 maintains the highest agreement (ICC = 0.883), followed closely by Claude V2 (0.865) and GLM V2 (0.853). Notably, both GPT variants converge at ICC = 0.820, indicating that rubric enhancement does not improve GPT's performance on the most complex questions that is a divergence from the consistent V1-to-V2 gains observed at L1--L3. This suggests that L4 questions, which require multi-concept integration and system-wide reasoning, present challenges that cannot be overcome solely through rubric specificity.

Across all models, the Global ICC reported in Table~\ref{tab:human_ai_agreement} falls between the L2 and L3 values, reflecting the composite effect of questions distributed across all four taxonomy levels. The consistent pattern (highest agreement at L1, moderate agreement at L2, and systematic degradation at L3--L4) establishes cognitive complexity as a reliable predictor of automated evaluation difficulty. This finding has direct practical implications: instructors can use the proposed taxonomy to stratify their question banks, delegating L1--L2 questions to AI-assisted grading with high confidence while reserving L3--L4 questions for human review or hybrid workflows.

The rubric effect is most pronounced at intermediate complexity levels (L2--L3), where the difference between V1 and V2 variants is largest for most models. At L1, some models (e.g., Claude) already perform well under the baseline prompt, leaving less room for improvement. At L4, rubric guidance improves performance but does not fully compensate for the inherent difficulty of evaluating multi-concept reasoning. This suggests that rubric quality is a necessary but insufficient condition for reliable AI-assisted grading of high-complexity questions, and that even the best-performing models require human oversight for L4 assessment.

The mean human evaluation scores provide a useful external benchmark against which the LLM-generated grades can be assessed. Human evaluators produced a mean final grade of 5.41 and a median of 5.70, figures that are closely aligned with the V2 configurations: GPT V2 (mean 5.40, median 5.70) and Gemini V2 (mean 5.39, median 5.85) which are notably higher than all V1 configurations. However, the distributional profiles diverge in several respects. Human evaluators assigned a Distinction rate of 5.3\%, substantially higher than any model in either variant (maximum 2.3\%), and a Merit rate of 25.29\%, comparable to the Merit rates achieved by GPT V2 (28.7\%) and Gemini V2 (28.4\%) but markedly higher than any V1 configuration. These observations indicate that while the V2 variants of GPT and Gemini approximate human central tendency reasonably well, no model fully replicates the distributional shape of human evaluation, particularly at the upper end of the grade spectrum.

The empirical results presented above carry several theoretical implications regarding the nature of automated assessment and its relationship to expert human evaluation. A first observation concerns the role of knowledge that experienced instructors apply intuitively, beyond what any rubric can capture. The persistent gap between LLM and human agreement, even under rubric-guided prompting, suggests that expert evaluators draw on contextual understanding that cannot be fully codified in a rubric. Instructors recognise pedagogical value in partial attempts, distinguish productive struggle from genuine misconception, and adjust credit accordingly. This judgments rely on accumulated teaching experience rather than explicit criteria. Current LLMs, trained to match surface-level textual patterns, do not replicate this interpretive layer.

A second implication concerns the relationship between rubric quality and evaluation reliability. The consistent improvement in human--LLM agreement from V1 to V2 across all four models demonstrates that the benefits of structured criteria extend beyond human evaluators to automated systems. When a rubric is vague or incomplete, an experienced instructor fills the gaps using accumulated pedagogical knowledge, for instance, recognising that a student understood the underlying concept even though their answer contained a syntax error. An LLM has no such background: those same gaps become points of failure, causing the model to grade incorrectly where a human would exercise informed judgment. Supplying the full rubric in V2 reduces this ambiguity, which accounts for the uniform improvement observed across all four models. This suggests that the discipline required to construct a precise, comprehensive rubric yields returns not only for human grading consistency but also for the viability of AI-assisted evaluation. Institutions considering automated grading should therefore treat rubric development as a prerequisite rather than an optional step.

A third and particularly salient limitation concerns cross-question context and intent recognition. Manual inspection of individual responses revealed a recurring failure mode in which LLMs penalised answers for using paths or filenames that did not match the model solution, even when the discrepancy arose because the student had correctly constructed those artefacts in a preceding question. In examinations structured as sequential tasks, where outputs of one question become inputs to the next, evaluating each item in isolation systematically misattributes coherent, consistent responses as incorrect. This limitation is not purely a prompt-engineering problem: while providing the full examination context in the prompt could partially mitigate the issue, it would not resolve the deeper challenge. Human evaluators do not merely verify surface-level answer accuracy, they interpret student intent and assess whether the response demonstrates the underlying knowledge, distinguishing productive approximations from genuine misconceptions. This capacity to separate \textit{what a student knows} from \textit{how precisely they performed on a given item} is a tacit competency that current LLMs lack, and it represents one of the most significant qualitative gaps between automated and expert human assessment in practice-oriented technical courses. The distinction is not merely technical but fundamentally pedagogical: the purpose of assessment is to measure learning, not to perform pattern matching against a reference solution. When an automated system assigns zero credit to an answer that is functionally coherent but textually divergent from the expected output, it does not produce a wrong grade by a small margin, it produces a categorically invalid grade, one that misrepresents the student's actual level of competence. This validity failure is especially consequential in sequential, task-based examinations, where a single surface-level discrepancy can cascade across multiple items and systematically distort the final grade of a student who has, in fact, demonstrated mastery. Human evaluators avoid this failure mode not because they apply a more sophisticated rubric, but because they engage in genuine interpretive reasoning: they ask what the student was trying to do, whether the reasoning behind the answer is sound, and whether the error is symptomatic of a conceptual gap or merely an artefact of notation.

Taken together, the quantitative agreement analysis and the theoretical considerations converge on a coherent picture: current frontier LLMs can serve as reliable assistants for automated grading at lower cognitive levels, provided that evaluation criteria are made explicit through well-structured rubrics, but they cannot yet substitute human judgment on questions that require structural reasoning, cross-item context, or the interpretation of student intent. The practical implication is not a binary choice between human and automated grading, but rather a principled allocation of effort guided by cognitive complexity. 
The following section summarises the main contributions of this work, discusses its limitations, and outlines directions for future research.

\section{Conclusion}
\label{sec:conclusion}

This work addresses the challenge of scalable, reliable assessment in technical computing courses by systematically evaluating LLM-assisted grading of bash command responses against expert human evaluation. A four-level cognitive taxonomy that integrates Bloom's Revised Taxonomy with an operational-impact dimension serves as a scaffolding, providing a principled criterion for stratifying question difficulty and predicting where automated evaluation is likely to succeed or fail. Building on this, human evaluation was designed using this four-level taxonomy, supported by a range of statistical measures that reinforce its validity. It also establishes a structured framework for comparing the performance of the four evaluated LLMs under baseline and rubric-guided prompting.

The empirical findings reveal a consistent pattern across all models: LLM--human agreement is highest at lower taxonomy levels (L1--L2), where questions target information retrieval and basic manipulation, but degrades systematically at higher levels (L3--L4), where structural reasoning about permission models, data streams, and multi-concept integration is required. This makes the taxonomy level a reliable predictor of automated evaluation difficulty. A key practical finding is that rubric quality matters more than model choice: well-specified rubrics consistently improve agreement across all models tested, yielding larger gains than switching between provider ecosystems. Consequently, institutions considering AI-assisted grading should invest in rubric development before selecting a specific model, and instructors should prioritize human review for complex questions while relying on AI-assisted grading for lower-level tasks where feedback is most reliable.

The systematic empirical effort undertaken in this study opens several concrete directions for future research. The most immediate extension concerns the generalizability of the evaluation framework to other technical assessment domains. Even if the taxonomy is tailored to the context of bash commands, it can be generalized to other domains, such as database query evaluation or network administration. Extending the underlying evaluation procedures and methodology to other domains would establish whether the relative performance of models and prompt variants is stable. 

Another direction concerns prompt engineering and model configuration. This study evaluated two prompt variants under default model settings, leaving unexplored a range of design choices that may substantially affect agreement with human evaluators. Few-shot prompting, or iterative self-refinement represent the most promising structural variants, while systematic exploration of model-level parameters such as temperature offers a complementary axis of investigation. Future work should also examine whether prompts optimized for one model family transfer to others, or whether effective prompt design is inherently model-specific.

The dataset will be expanded in future examination sittings of the same course, incorporating additional questions and a broader range of assessment formats, all classified according to the  taxonomy. This progressive growth will provide a richer and more diverse evidence base, enabling more stable statistical conclusions and more granular analyses across taxonomy levels and question types than the current sample supports.

\section*{Acknowledgments}
Funding for open access charge: Universidade de Vigo/CISUG.

\section*{Data availability}
The data used in this study are available from the corresponding author on reasonable request.

\bibliographystyle{elsarticle-harv}
\bibliography{refs}

\end{document}